\documentclass[lettersize,journal]{IEEEtran}
\usepackage{amsmath,amsfonts}
\usepackage{algorithm}
\usepackage{array}
\usepackage[caption=false,font=normalsize,labelfont=sf,textfont=sf]{subfig}
\usepackage{textcomp}
\usepackage{stfloats}
\usepackage{url}
\usepackage[colorlinks=true,linkcolor=blue,citecolor=blue,urlcolor=blue]{hyperref} 
\usepackage{verbatim}
\usepackage{graphicx}
\usepackage{cite}
\usepackage{multirow}
\usepackage{algpseudocode}
\usepackage{siunitx}
\usepackage{booktabs}
\usepackage{makecell}
\usepackage{orcidlink}
\usepackage{array}
\usepackage{subcaption}
\begin{document}

\title{Enhancing Implicit Neural Representations with Image Feature Embedding for Unsupervised Cardiac Cine MRI Reconstruction}

\author{Donghang Lyu\orcidlink{0009-0008-1446-9930}, Marius Staring\orcidlink{0000-0003-2885-5812}, Yiming Dong\orcidlink{0000-0002-8580-4555}, Keupp Jochen\orcidlink{0000-0003-2129-1153}, Hildo J. Lamb\orcidlink{0000-0002-6969-2418}, Mariya Doneva\orcidlink{0000-0003-0292-5069}
\thanks{Manuscript received June XX, 2026. This work was supported by
the project ROBUST: Trustworthy
AI-based Systems for Sustainable Growth with project number
KICH3.LTP.20.006. (corresponding author:
Donghang Lyu)}
\thanks{Donghang Lyu, and Marius Staring are with the
Division of Image Processing, Department of Radiology, Leiden University
Medical Center, 2333ZA Leiden, Netherlands. (e-mail: d.lyu@lumc.nl;
m.staring@lumc.nl)}
\thanks{Yiming Dong is with the C.J. Gorter
Center, Department of Radiology, Leiden University
Medical Center, 2333ZA Leiden, Netherlands. (e-mail: y.dong1@lumc.nl)}
\thanks{Hildo J. Lamb is with the Cardio Vascular Imaging Group, Department of Radiology, Leiden University
Medical Center, 2333ZA Leiden, Netherlands. (e-mail: h.j.lamb@lumc.nl)}
\thanks{Keupp Jochen, and Mariya Doneva are with
Philips Innovative Technologies, 22335 Hamburg, Germany. (e-mail: jochen.keupp@philips.com; mariya.doneva@philips.com)}} 

\markboth{2026}%
{Shell \MakeLowercase{\textit{et al.}}: A Sample Article Using IEEEtran.cls for IEEE Journals}

\maketitle


\begin{abstract}
Cardiac cine Magnetic Resonance Imaging (MRI) is a critical diagnostic tool that provides dynamic insights for radiologists. To accelerate acquisition, under-sampled k-space data is often used, requiring reconstruction methods that combine coil sensitivity encoding with prior information to recover missing data. Deep learning approaches have gained more attention for leveraging data-adaptive priors. While supervised learning approaches are a common choice, they depend on fully sampled reference data, which is not always available. Unsupervised methods eliminate the need for fully sampled reference data, which can be advantageous in cardiac cine MRI reconstruction. Among them, implicit neural representations (INRs) have shown great potential due to their simple architecture and good quality reconstructions. In this work, we propose an image-domain dual-branch INR framework, termed I-FP-INR, which extends the original INR design by introducing an additional feature-processing branch. This design aims to extract complementary feature embeddings to enhance the overall representation, thereby benefiting reconstruction. Extensive evaluations on both public datasets and in-house data show consistent improvements over baseline methods in reconstruction quality, with strong robustness across varied scenarios.
\end{abstract}


\section{Introduction}
\IEEEPARstart{C}{ardiac} cine Magnetic Resonance Imaging (MRI) enables dynamic assessment of cardiovascular function by providing excellent soft-tissue contrast and capturing cardiac motion, thereby serving as an essential modality for clinical diagnosis. However, patients must remain still during acquisition, and long scan times with repeated breath-holds can cause discomfort and limit high-resolution imaging~\cite{1}. To overcome these challenges, higher acceleration factors are used to shorten scans and reduce breath-hold demands, requiring advanced reconstruction methods to recover high-quality images from highly under-sampled data.

Parallel imaging~\cite{2,42} and compressed sensing~\cite{43} are widely used to reconstruct MR images from under-sampled k-space data~\cite{3}. Parallel imaging exploits multi-coil sensitivity encoding, while compressed sensing relies on sparsity to recover missing data. However, parallel imaging with 1D under-sampling is typically limited to acceleration factors of around 4~\cite{44} due to noise amplification, whereas compressed sensing with non-uniform sampling can achieve slightly higher acceleration. For dynamic acquisitions such as cardiac cine, additional gains can be achieved by incorporating temporal correlations across frames. In recent years, deep learning methods have gained increasing attention for their stronger representation capability and reconstruction performance. Numerous supervised models~\cite{4,5,6,7,8} have been proposed to learn mappings from under-sampled to fully sampled data, with unrolled networks showing strong performance by combining iterative reconstruction with learned regularization.  In the context of cine MRI, the additional temporal dimension and strong spatio-temporal correlations within images have further motivated specialized approaches~\cite{9,10,11}, leading to promising results for highly accelerated reconstruction.

Although supervised learning approaches have demonstrated impressive MRI reconstruction performance, acquiring a large number of fully sampled images remains challenging in many practical scenarios. In contrast, self-supervised learning methods~\cite{12,13,14,15} address this by removing the need for fully sampled references, though they may lag in performance and require longer reconstruction times. The SSDU framework~\cite{12} splits acquired k-space measurements into two disjoint subsets, using one subset for training and the other for data consistency. Subsequent extensions~\cite{13,17} further improve this framework by incorporating Noise2Noise~\cite{16} or weighted partition strategies. However, these approaches still depend on large-scale datasets. To reduce this dependency, some alternative methods~\cite{15,19} adopt the Deep Image Prior (DIP) framework~\cite{18}, which optimizes a randomly initialized network per scan to generate images consistent with measured k-space data via the forward model, but typically requires long reconstruction times.

Beyond the aforementioned methods, Implicit Neural Representations (INRs) offer a promising self-supervised reconstruction paradigm by modeling signals as continuous functions, eliminating the need for large datasets. Accordingly, a multi-layer perceptron (MLP) is typically used to learn a continuous mapping from spatial coordinates to  signal intensities~\cite{20}. In medical imaging, INRs have been applied to various tasks such as registration~\cite{22}, segmentation~\cite{23}, growth prediction~\cite{45} and image reconstruction~\cite{14}, highlighting its strong potential.

Some recent works~\cite{14,24,25} have explored INR-based methods for cine MRI reconstruction, demonstrating good image quality at high acceleration factors. These methods fall into two categories based on whether intensity estimation is performed in k-space or image space. In k-space approaches, NIK~\cite{14} is a representative work that employs an MLP to map inputs (time point, k-space coordinates, and coil index) from the sampled region to k-space values, thereby learning the full k-space distribution and estimating missing data. Building on this, PISCO~\cite{26} introduces an additional regularization term to encourage the learning of local neighborhood relationship in k-space. In contrast, image-domain INR methods take the full spatial coordinate grid as input to predict image intensities, which are then transformed into k-space via fast Fourier transform (FFT) for enforcing k-space consistency. Among these, FMLP~\cite{25} separately embeds the spatial and temporal components, followed by an MLP to predict complex-valued image intensities. Hash-INR~\cite{24} combines an MLP with hash encoding and applies k-space data consistency loss along with image-domain regularization.

While INR methods have shown promising results for cine MRI reconstruction in both k-space and image domains, most approaches still follow the conventional paradigm of predicting target values solely from coordinates. Beyond spatial coordinates, incorporating feature embeddings can further enhance performance by enriching each target point with contextual information from its neighborhood. For instance, IF-Nets~\cite{28} construct a multi-scale 3D feature grid that captures both global and local shape properties, which are sampled using spatial coordinates and decoded for accurate 3D reconstruction. Similarly, KP-INR~\cite{27} firstly explores k-space feature embeddings by using a feature extractor to capture spatio-temporal context, followed by an additional INR branch that interacts with the coordinate-based representation. Since the initial k-space input is undersampled and provides limited context, an inference stage is introduced and alternates with the optimization process to progressively refine the k-space input. This iterative refinement improves the quality of k-space feature embeddings by leveraging increasingly accurate neighborhood estimates, thereby leading to more effective learning. Overall, KP-INR outperforms the original INR design without k-space feature embeddings, demonstrating the feasibility and effectiveness of integrating feature embeddings into INR-based cine MRI reconstruction.

Although INR-based methods in k-space have shown promise, they still exhibit several limitations. First, they rely solely on sampled coordinates during optimization, making it challenging to capture the global k-space distribution at high acceleration factors. Then, since each k-space point influences all image pixels, misaligned k-space values can easily introduce pronounced artifacts in the reconstructed images~\cite{25}. Moreover, the highly non-uniform SNR distribution in k-space, dominated by low-frequencies, further complicates learning of INR. As a result, KP-INR also remains affected by these inherent drawbacks. To address this, we extend our prior work KP-INR to the image domain, termed I-FP-INR, avoiding the aforementioned limitations. Instead of k-space inputs, I-FP-INR uses coordinate grids and zero-filled images, with both branches producing reconstructed images that are transformed back to k-space for data consistency optimization. Since zero-filled reconstructions contain aliasing artifacts, the extracted features may be suboptimal and provide less informative complementary representations. Therefore, we adopt the KP-INR design by introducing an inference stage that iteratively refines the image inputs, leading to a more stable optimization process and more accurate feature representations. We also explore I-FP-INR under multi-coil and coil-combined reconstruction strategies, showing that each has distinct advantages depending on the sampling pattern setting. The key contributions of this work are as follows:
\begin{itemize}
    \item We propose I-FP-INR, which enhances coordinate-based representations by incorporating extracted image feature embeddings. The model introduces two interactive INR branches along with an alternating inference refinement stage within the optimization process, leading to improved performance in cardiac cine MRI reconstruction compared to baseline methods.
    \item Through extensive ablation studies, the design of I-FP-INR is validated. We further investigate multi-coil and coil-combined reconstruction strategies across diverse sampling patterns, showing that each offers advantages under different scenarios.
    \item We evaluate I-FP-INR on the public CMRxRecon2024 dataset~\cite{29} across two acceleration factors of 4$\times$ and 8$\times$ under multiple sampling patterns, as well as on a private in-house data with varying acceleration factors and sampling masks, demonstrating robust and consistent reconstruction performance.
\end{itemize}


\section{Problem Formulation}

Using an INR method for cardiac cine MRI reconstruction in the image domain can be formulated as learning an implicit image representation by enforcing consistency with a forward model that maps the reconstructed images to the measured k-space data. Let the reconstructed images be denoted as $\textbf{x} \in \mathbb{C}^{N \times T}$, and the measured k-space data from the $c$-th coil at frame $t$ as $\textbf{y}_{c}^{t} \in \mathbb{C}^{M_t} (1 \le c \le C, 1 \le t \le T)$. Here, $N=D_{h} \times D_{w}$ represents the total number of spatial pixels, $T$ is the number of temporal frames, $M_{t}$ denotes the number of sampled k-space points at frame $t$, and $C$ is the coil number. The forward model is then given by:
\begin{equation}
    \textbf{y}_{c}^{t} = M_{u}^{t}\mathcal{F}S_{c}\textbf{x}^{t}.
\end{equation}
$S_{c} \in \mathbb{C}^{N \times N}$ is a diagonal matrix representing
the $c$-th coil sensitivity map, $\mathcal{F}$ represents the Fourier transform, and $M_{u}^{t} \in \mathbb{R}^{M_{t} \times N}$ denotes the undersampling operator at frame $t$ that models the acquisition mask.

Reconstructing the image $\textbf{x}$ from undersampled k-space measurements constitutes an ill-posed inverse problem, which can be formulated as the following optimization problem:
\begin{equation} \label{eq2}
\arg\min_{\textbf{x}} \; \sum_{t=1}^{T}\sum_{c=1}^{C}
\left\| M_{u}^{t} \mathcal{F} S_c \textbf{x}^{t} - \textbf{y}_{c}^{t} \right\|_2^2 + \mathcal{R}(\textbf{x}),
\end{equation}
where $\mathcal{R}$ is a regularization term that promotes desirable image properties and stabilizes the reconstruction from undersampled data. In compressed sensing~\cite{30,31}, hand-crafted priors such as low-rank constraints or Total Variation (TV) are commonly used. In contrast, deep learning-based approaches replace these with neural networks that learn data-driven priors, improving cine MRI reconstruction quality.

In the context of INR-based reconstruction, the core idea is to learn a continuous mapping function $f_{\theta}$ that represents the data, enabling compact representation and generation beyond discrete samples. Given a coordinate grid $P \in \mathbb{R}^{N\times3}$, the function maps each coordinate $p=(i, j, t)$ to its corresponding image intensity value $\mathbf{x}_{p}$. Here, $(i,j)$ is the 2D spatial coordinate with $1\le i \le D_{h}$ and $1 \le j \le D_{w}$, while $t$ denotes the temporal coordinate with $1 \le t \le T$. Accordingly, the reconstructed image sequence $\textbf{x}$ can be parameterized by $\theta$ as $\textbf{x}_{\theta}$, and the overall process is formulated as:
\begin{equation} \label{eq4}
    \arg\min_{\theta} \; \sum_{t=1}^{T}\sum_{c=1}^{C}
    \left\| M_{u}^{t} \mathcal{F} S_c \textbf{x}_{\theta}^{t} - \textbf{y}_{c}^{t} \right\|_2^2 + \lambda \left\| \textbf{x}_{\theta}-f_{\theta}(P) \right\|^2,
\end{equation}
where $\lambda$ is a hyper-parameter controlling the strength of the regularization term.

\section{Method}
\subsection{Overview of I-FP-INR}
\begin{figure*}[!t]
\centering
\subfloat[The overall optimization pipeline of I-FP-INR with a multi-coil reconstruction strategy.]{\includegraphics[width=5.6in]{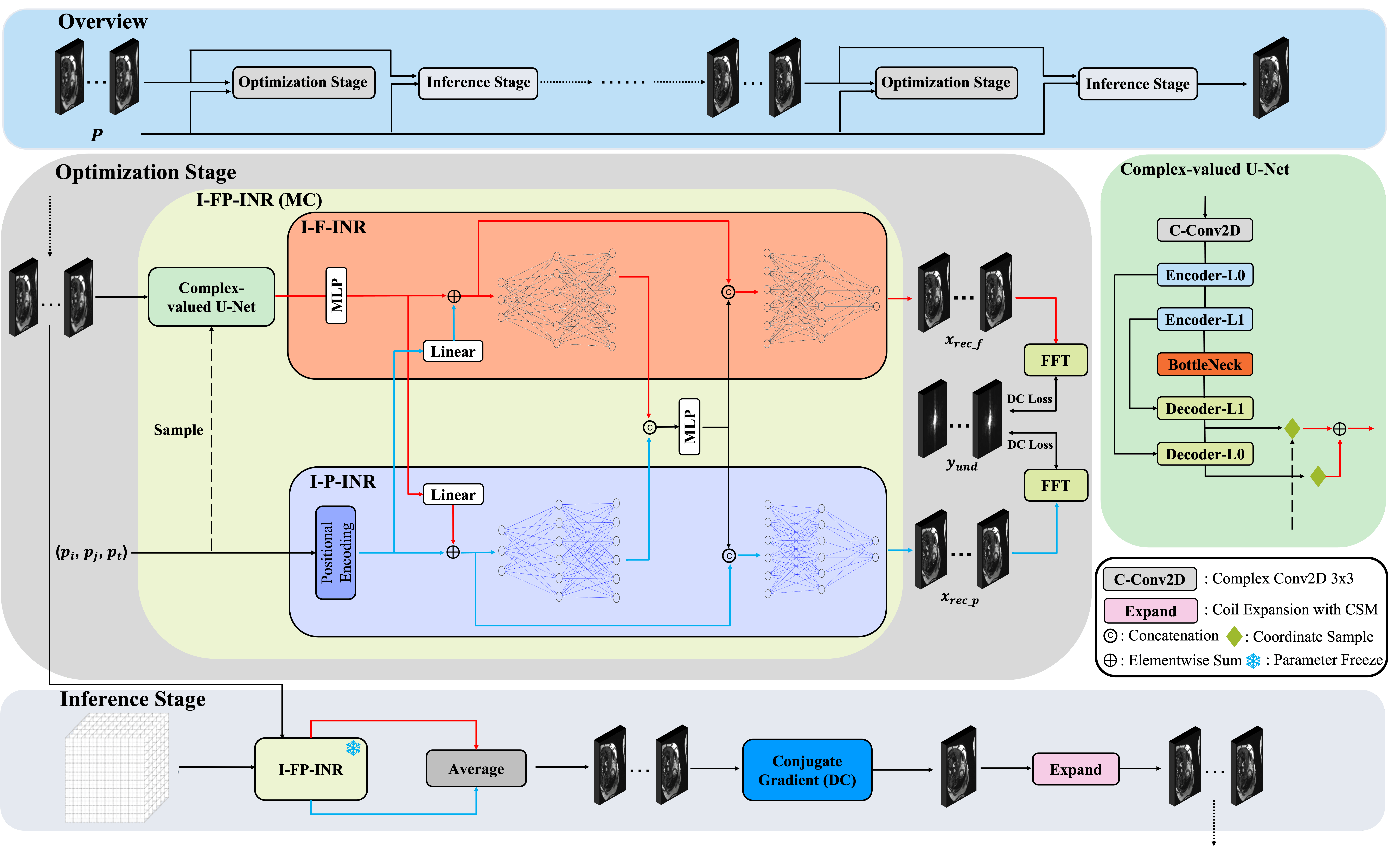}%
\label{fig1_mc}}
\hfil
\subfloat[The overall optimization pipeline of I-FP-INR with a coil-combined reconstruction strategy.]{\includegraphics[width=5.6in]{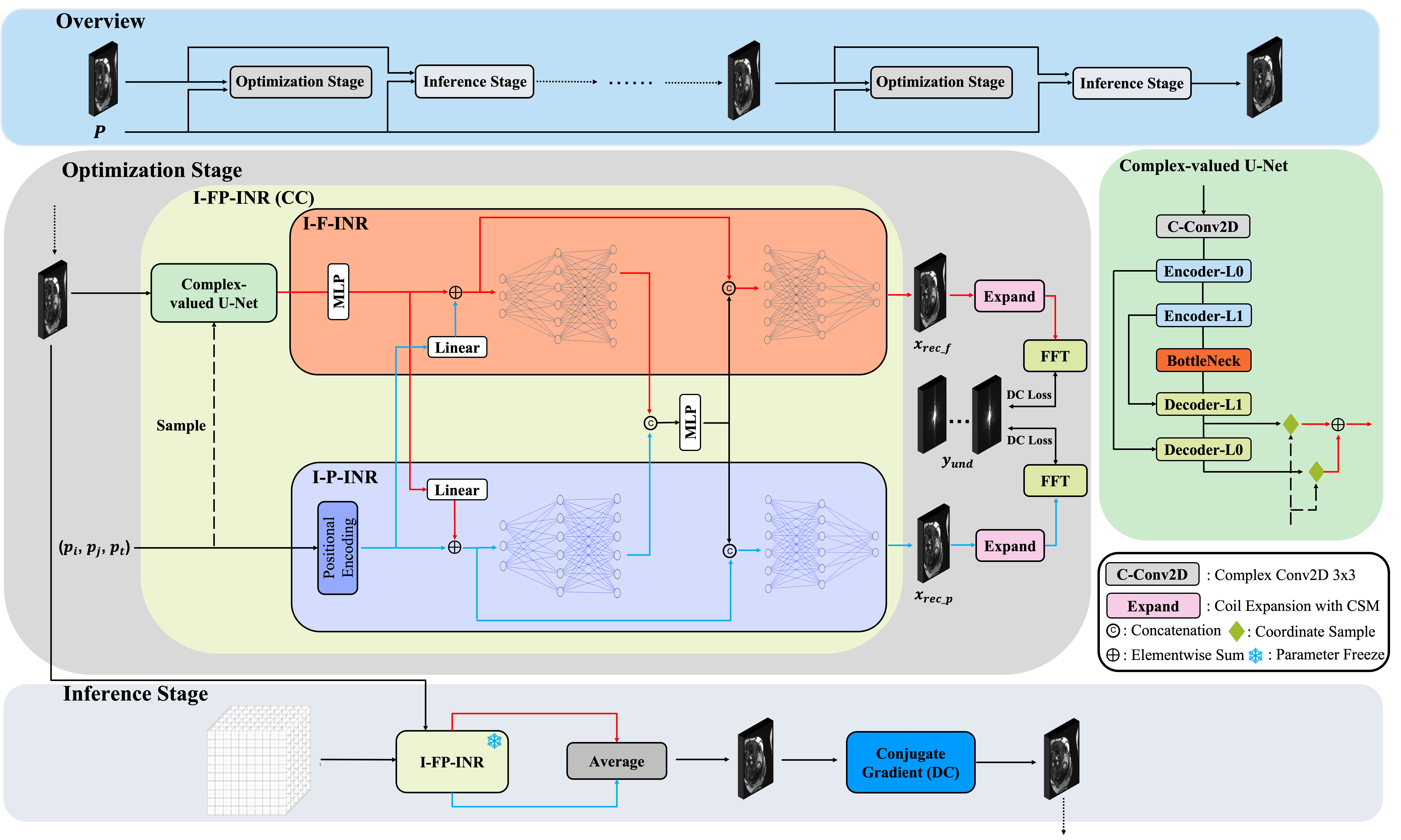}%
\label{fig1_sc}}
\caption{The overall optimization pipeline of I-FP-INR under two reconstruction strategies is illustrated: (a) the multi-coil strategy, and (b) the coil-combined strategy. In each sub-figure, the top panel depicts the iterative image updates using the I-FP-INR method with the coordinate grid $P$, where dotted lines indicate the propagation of the refined cine sequence. Within the model, blue and red lines represent the flow of positional embedding and image feature embedding, respectively. The bottom panel shows the inference pipeline, and the right panel illustrates the overall architecture of the complex-valued U-Net with the coordinate-based feature sampling process.}
\label{fig1}
\end{figure*}

As shown in Fig.~\ref{fig1}, I-FP-INR comprises two branches. The I-P-INR branch follows the original formulation in Eq.~\ref{eq4}, learning the mapping between spatial coordinates and their corresponding values. In contrast, the I-F-INR branch takes feature embeddings extracted from cine images $x$ via a complex-valued U-Net as input, providing complementary feature information.

Therefore, for I-FP-INR, Eq.~\ref{eq4} can be reformulated as:
\begin{equation}\label{eq6}
        \arg\min_{\theta} \; \sum_{t=1}^{T}\sum_{c=1}^{C}
    \left\| M_{u}^{t} \mathcal{F} S_c \textbf{x}_{\theta}^{t} - \textbf{y}_{c}^{t} \right\|_2^2 + \lambda \left\| \textbf{x}_{\theta}-f_{\theta}(P, x) \right\|^2.
\end{equation}
For the I-P-INR branch, the input coordinate grid $P$ is first mapped to a high-dimensional embedding using positional encoding. Following~\cite{25}, we decompose each coordinate into spatial and temporal components and encode them separately. This design enhances the recovery of high-frequency details, leading to sharper reconstructions. Specifically, we choose to adopt neural radiance field (NeRF) positional encoding $\gamma_{s}$~\cite{32} for the spatial component $p_{s}=(i,j)$, and Fourier positional encoding $\gamma_{t}$~\cite{33} for the temporal component $p_{t}=t$. The resulting representations are concatenated to form the final $F_p$-dimensional positional embedding $\varphi_{p} \in \mathbb{R}^{F_p}$, where $p$ indexes each coordinate. The complete positional encoding process can be written as:
\begin{equation}
\begin{aligned}
\varphi_{s} = \gamma_s(p_s) = [&\, \sin(2^0 \pi p_s), \cos(2^0 \pi p_s), \\
                 &\, \ldots, \\
                 &\, \sin(2^{L_s-1} \pi p_s), \cos(2^{L_s-1} \pi p_s) \,],
\end{aligned}
\end{equation}
\begin{equation}
 \varphi_{t}  = \gamma_t(p_t) = 
\big[
\sin(\mathbf{B} p_t), \cos(\mathbf{B} p_t)
\big],
\end{equation}
\begin{equation}
    \varphi_{p} = \text{concat}(\varphi_{s}, \varphi_{t}).
\end{equation}
Here, $\varphi_{s}$ and $\varphi_{t}$ denote the spatial and temporal positional embeddings, respectively. A random frequency vector $\mathbf{B} \sim \mathcal{N}(0, \sigma^2)$ is sampled from a Gaussian distribution. The resulting positional embedding is then passed through an MLP network with complex Gabor wavelet activations (WIRE)~\cite{34}, which enhances the model’s ability to capture both fine details and global structure, producing the reconstructed image sequence $\textbf{x}_{\text{rec}\_\text{p}}$.

For the I-F-INR branch, the MLP operates on image feature embeddings instead of raw coordinate representations, requiring a feature extractor beforehand. Following~\cite{27}, we adopt a lightweight complex-valued U-Net suited for complex-valued inputs, preserving both phase and magnitude information while aggregating local context. The architecture is based on CRUNet~\cite{10}, which effectively exploits spatio-temporal information within frames. Specifically, we incorporate two complex-valued convolutional recurrent layers with opposite information propagation directions into the encoder and decoder, respectively. No downsampling is applied to avoid information loss and enable direct coordinate-based feature sampling from the decoder part. We also simplify the structure by removing side branches at each level. Considering the two-level design, this yields two $F_f$-dimensional feature embeddings, $\varphi_{f1} \in \mathbb{R}^{F_f}$ and $\varphi_{f2} \in \mathbb{R}^{F_f}$, are obtained, which are then fused via a learnable weighting parameter:
\begin{equation}
    \varphi_{f} = \alpha \cdot \varphi_{f1} + (1-\alpha) \cdot \varphi_{f2},
\end{equation}
where $\alpha \in [0,1] $ is a learnable scalar parameter. The two feature embeddings are fused using this weight and then passed through an MLP to match the dimensionality of the positional embeddings. The resulting representation is fed into the INR branch with Leaky ReLU activations to generate the reconstructed output $\textbf{x}_{\text{rec}\_\text{f}}$, enabling complementary feature learning to enrich the overall representation.

To facilitate effective information exchange between the two INR branches, we introduce two interaction modules: one before the processing of INR branches and another embedded within them. In the first module, separate linear layers are applied to each embedding, allowing them to incorporate information from each other at the initial stage:
\begin{equation}
    \varphi_{f}' = \varphi_{f}+\text{linear}(\varphi_{p}),
\end{equation}
\begin{equation}
    \varphi_{p}' = \varphi_{p}+\text{linear}(\varphi_{f}).
\end{equation}
In the second module, the embeddings are further integrated midway through each INR branch. The resulting embedding is then concatenated with the initial input embedding via a skip connection, preserving the original information while incorporating cross-branch features. The overall information fusion process can thus be summarized as follows:
\begin{equation}
    \varphi_{\text{fuse}} = \text{concat}(\text{MLP}_{\theta_{f}}(\varphi_{f}'), \text{MLP}_{\theta_{p}}(\varphi_{p}')),
\end{equation}
\begin{equation}
    \varphi_{f}' = \text{concat}(\varphi_{f}', \varphi_{\text{fuse}}),
\end{equation}
\begin{equation}
   \varphi_{p}' = \text{concat}(\varphi_{p}', \varphi_{\text{fuse}}).
\end{equation}

To enable effective learning in both INR branches and consistency with measured data, we apply a data consistency loss on the measured k-space regions, encouraging accurate reconstructions from both branches. An inference-stage refinement is then introduced after a certain number of optimization iterations. The image input and coordinate grid are passed through the frozen I-FP-INR model, and the two outputs are averaged to produce a preliminary reconstruction. Data consistency is subsequently enforced by using the conjugate gradient (CG) algorithm, yielding the final refined image. This refinement provides improved inputs for the I-F-INR branch, enabling more accurate feature extractions and further enhancing overall learning. 

\subsection{I-FP-INR with Two Reconstruction Strategies}
As mentioned earlier, I-FP-INR is investigated with two reconstruction strategies: multi-coil and coil-combined, leading to slight variations in image input, branch outputs and the inference stage, as illustrated in Fig.~\ref{fig1_mc} and Fig.~\ref{fig1_sc}.

The coil-combined reconstruction strategy simplifies learning by operating in a reduced-dimensional image space, improving stability and efficiency. Rather than directly modeling multi-coil k-space data, the network learns a continuous representation from coil-combined images, with coil-specific information reintroduced via physical constraints. During optimization, coil-combined images are used as input to produce an image consistent with the continuous representation, which is then expanded into coil-wise images via coil sensitivity maps (CSMs) and transformed to k-space for data consistency enforcement. During inference, the two coil-combined outputs are averaged to obtain an INR estimate, which is further refined by a CG-based data consistency module. The refined image is then used as input for the next optimization stage. More details are summarized in the algorithm block below.

\begin{algorithm}[!tbh]
\caption{Two-Stage I-FP-INR Optimization Pipeline with Coil-Combined Reconstruction Strategy}
\label{alg:sampling2}
\begin{algorithmic}[1]
\State \textbf{Input:} A Cartesian grid including all coordinates $\mathcal{P}_{grid}$, total number of optimization steps $T$, interval for inference $t$, sampling mask $M$, coil sensitivity map $S$, conjugated coil sensitivity map $S^{-1}$, Fourier transform $\mathcal{F}$, inverse Fourier transform $\mathcal{F}^{-1}$, the undersampled multi-coil k-space $\mathbf{y}_{\text{und}}^{\text{mc}}$ 
\For {$i=0, \dots, T$}
    \State Optimization Stage:
    \If{$i \le t$}
    \State $\mathbf{x}^{\text{cc}} \gets \sum S^{-1} \odot \mathcal{F}^{-1}(\mathbf{y}_{\text{und}}^{\text{mc}})$
    \EndIf
    \State $\mathbf{x}_{\text{rec\_f}}^{\text{cc}},\ \mathbf{x}_{\text{rec\_p}}^{\text{cc}} \gets \texttt{I-FP-INR}_{\theta}(\mathcal{P}_{grid},\ \mathbf{x}^{\text{cc}})$
    \State $\mathbf{y}_{\text{rec\_f}}^{\text{mc}}, \mathbf{y}_{\text{rec\_p}}^{\text{mc}} \gets \mathcal{F}(S \odot \mathbf{x}_{\text{rec\_f}}^{\text{cc}}), \mathcal{F}(S \odot \mathbf{x}_{\text{rec\_p}}^{\text{cc}})$
  \State Calculate loss: $\mathcal{L} = \mathcal{L}_{\mathbf{y}_{\text{rec\_f}}^{\text{mc}}} + \mathcal{L}_{\mathbf{y}_{\text{rec\_p}}^{\text{mc}}}$
  \Comment{Data consistency loss in the sampled region of multi-coil k-space}
  \State Optimizing model parameters: $\theta \leftarrow \theta - \alpha \nabla_\theta \mathcal{L}$
    \State Inference Stage:
    \If{$i \bmod t = 0 \land i > 0$}
        \State $\mathbf{x}_{\text{rec\_f}}^{\text{cc}},\ \mathbf{x}_{\text{rec\_p}}^{\text{cc}} \gets \texttt{I-FP-INR}_{\theta}(\mathcal{P}_{\text{grid}},\ \mathbf{x}^{\text{cc}})$
        \State $\mathbf{x}_{\text{rec}}^{\text{cc}} \gets 0.5 \cdot \left( \mathbf{x}_{\text{rec\_f}}^{\text{cc}} + \mathbf{x}_{\text{rec\_p}}^{\text{cc}} \right)$
        
        \State Conjugate Gradient Data Consistency:
        \State $\mathbf{x}_{\text{rec}} \gets \texttt{CG}(\mathbf{x}_{\text{rec}}^{\text{cc}}$, $\mathbf{y}_{\text{und}}^{\text{mc}}$, $M$, $S)$ 
        \Comment{Get coil-combined reconstructed output $\mathbf{x}_{\text{rec}}
    = \left( \mathcal{A}^H \mathcal{A} + \lambda \mathcal{I} \right)^{-1}
    \left( \mathcal{A}^H(\mathbf{\mathbf{y}_{\text{und}}^{\text{mc}}}) + \lambda \mathbf{x}_{\text{rec}}^{\text{cc}}\right)$, forward operator $ \mathcal{A} = M \mathcal{F} S$}

    \State $\mathbf{x}^{\text{cc}} \gets \mathbf{x}_{\text{rec}}$
    \Comment{Updated coil-combined image input for the next optimization stage}

    \EndIf
\EndFor
\State \textbf{Output:} $\mathbf{x}_{\text{rec}}$
\end{algorithmic}
\end{algorithm}

Although the coil-combined image representation strategy is widely used in deep learning-based MRI reconstruction and aligns with Eq.~\ref{eq2}, modeling multi-coil images as channel-wise inputs might enable the network to explicitly learn inter-coil correlations and sensitivity encoding, potentially benefiting representation learning under certain scenarios. Importantly, this does not change the underlying reconstruction formulation: the optimization variable remains the coil-combined image series $\textbf{x}$, while the regularization term can be extended to operate in the multi-coil image space, i.e., $\mathcal{R}\left(S_c \mathbf{x} \right)$:
\begin{equation} \label{eq16}
\arg\min_{\mathbf{x}} \;
 \sum_{t=1}^{T}\sum_{c=1}^{C}
\left\| M_u^t \mathcal{F}(S_c \mathbf{x}^t) - \mathbf{y}_c^t \right\|_2^2
+ \mathcal{R}\left( S_c \mathbf{x} \right)
\end{equation}
As illustrated in Fig.~\ref{fig1_mc}, when multi-coil images are used as input, the model jointly estimates all coil images, which are transformed to k-space to compute the data-consistency loss over the sampled regions. The overall inference pipeline remains similar to the coil-combined one, with the main difference lying in the data-consistency step, as detailed in the following algorithm block, where outputs are expanded to the multi-coil representation for the next optimization stage.
\begin{algorithm}[!tbh]
\caption{Two-Stage I-FP-INR Optimization Pipeline with Multi-Coil Reconstruction Strategy}
\label{alg:sampling1}
\begin{algorithmic}[1]
\State \textbf{Input:} A Cartesian grid including all coordinates $\mathcal{P}_{grid}$, total number of optimization steps $T$, interval for inference $t$, sampling mask $M$, coil sensitivity map $S$, conjugated coil sensitivity map $S^{-1}$, Fourier transform $\mathcal{F}$, inverse Fourier transform $\mathcal{F}^{-1}$, the undersampled multi-coil k-space $\mathbf{y}_{\text{und}}^{\text{mc}}$ 
\For {$i=0, \dots, T$}
    \State Optimization Stage:
    \If{$i \le t$}
    \State $\mathbf{x}^{\text{mc}} \gets \mathcal{F}^{-1}(\mathbf{y}_{\text{und}}^{\text{mc}})$
    \EndIf
    \State $\mathbf{x}_{\text{rec\_f}}^{\text{mc}},\ \mathbf{x}_{\text{rec\_p}}^{\text{mc}} \gets \texttt{I-FP-INR}_{\theta}(\mathcal{P}_{grid},\ \mathbf{x}^{\text{mc}})$
    \State $\mathbf{y}_{\text{rec\_f}}^{\text{mc}}, \mathbf{y}_{\text{rec\_p}}^{\text{mc}} \gets \mathcal{F}(\mathbf{x}_{\text{rec\_f}}^{\text{mc}}), \mathcal{F}(\mathbf{x}_{\text{rec\_p}}^{\text{mc}})$
  \State Calculate loss: $\mathcal{L} = \mathcal{L}_{\mathbf{y}_{\text{rec\_f}}^{\text{mc}}} + \mathcal{L}_{\mathbf{y}_{\text{rec\_p}}^{\text{mc}}}$
  \Comment{Data consistency loss over the sampled region of multi-coil k-space}
  \State Optimizing model parameters: $\theta \leftarrow \theta - \alpha \nabla_\theta \mathcal{L}$
    \State Inference Stage:
    \If{$i \bmod t = 0 \land i > 0$}
        \State $\mathbf{x}_{\text{rec\_f}}^{\text{mc}},\ \mathbf{x}_{\text{rec\_p}}^{\text{mc}} \gets \texttt{I-FP-INR}_{\theta}(\mathcal{P}_{\text{grid}},\ \mathbf{x}^{\text{mc}})$
        \State $\mathbf{x}_{\text{rec}}^{\text{mc}} \gets 0.5 \cdot \left( \mathbf{x}_{\text{rec\_f}}^{\text{mc}} + \mathbf{x}_{\text{rec\_p}}^{\text{mc}} \right)$
        
        \State Conjugate Gradient Data Consistency:
        \State $\mathbf{x}_{\text{rec}} \gets \texttt{CG}(\mathbf{x}_{\text{rec}}^{\text{mc}}$, $\mathbf{y}_{\text{und}}^{\text{mc}}$, $M$, $S)$ 
        \Comment{Get coil-combined reconstructed output $\mathbf{x}_{\text{rec}}
    = \left( \mathcal{A}^H \mathcal{A} + \lambda \mathcal{I} \right)^{-1}
    \left( \mathcal{A}^H(\mathbf{\mathbf{y}_{\text{und}}^{\text{mc}}}) + \lambda \sum S^{-1} \odot \mathbf{x}_{\text{rec}}^{\text{mc}}\right)$, forward operator $ \mathcal{A} = M \mathcal{F} S$}
        
    \State $\mathbf{x}^{\text{mc}} \gets S \odot \mathbf{x}_{\text{rec}}$
    \Comment{Updated mutli-coil image input for the next optimization stage}

    \EndIf
\EndFor
\State \textbf{Output:} $\mathbf{x}_{\text{rec}}$
\end{algorithmic}
\end{algorithm}

\subsection{Loss Function}
As a self-supervised method, optimization relies solely on the acquired k-space measurements, with the loss computed on the measured samples. To address the large magnitude disparity between central low-frequency and peripheral high-frequency components, we combine the NIK loss~\cite{14} and relative L2 loss~\cite{16}. Both mitigate dynamic range imbalance: NIK loss combines a high dynamic range (HDR) reconstruction term with an explicit k-space regularization term, while relative L2 loss normalizes by the estimated real and imaginary components, balancing gradients and improving high-frequency reconstruction. The overall objective is formulated as:
\begin{equation}
    \mathcal{L}_{\text{NIK}}
=
\frac{1}{N}
\sum_{n=1}^{N}
\mathcal{L}_{\mathrm{HDR}}\bigl(\hat{\textbf{y}}_{n}, \textbf{y}_n\bigr)
+
\lambda_{\mathrm{r}}\, R\!\left(\hat{\textbf{y}}_{n}\right),
\end{equation}
\begin{equation}
\mathcal{L}_{\mathrm{HDR}}(\hat{\textbf{y}}, \textbf{y})
=
\left\|
\frac{\left|\hat{\textbf{y}} - \textbf{y}\right|}
{\left|\mathrm{sg}(\hat{\textbf{y}})\right| + \epsilon}
\right\|_2^2,
\end{equation}
\begin{equation}
    R\!\left(\hat{\textbf{y}}\right)
=
\left\|
\frac{\left|\hat{\textbf{y}} - \textbf{K}\hat{\textbf{y}}\right|}
{\left|\mathrm{sg}(\hat{\textbf{y}})\right| + \epsilon}
\right\|_2^2.
\end{equation}

\begin{equation}
\mathcal{L}_{\text{RelL2}}
= \frac{1}{N}
\sum_{n=1}^{N} \left\|
\frac{\hat{\textbf{y}}_n^{\text{real}} - \textbf{y}_n^{\text{real}}}
{\mathrm{sg}(\hat{\textbf{y}}_n^{\text{real}} ) + \epsilon}
\right\|_2^2 +
\left\|
\frac{\hat{\textbf{y}}_n^{\text{imag}} - \textbf{y}_n^{\text{imag}}}
{\mathrm{sg}(\hat{\textbf{y}}_n^{\text{imag}}) + \epsilon}
\right\|_2^2.
\end{equation}

Here, $\mathrm{sg}(\cdot)$ denotes the stop-gradient operator, and $|\cdot|$ computes magnitude. The constant $\epsilon$ avoids division by zero, while $\lambda_{\mathrm{r}}$ controls the contribution of the regularization term, which is 0.5. The weighting matrix $\mathbf{K} = e^{-d/2\sigma^{2}}$ is a distance-based weighting matrix, where $d = \sqrt{k_i^2 + k_j^2}$ denotes the radial distance to the k-space center, thereby modulating the regularization strength across k-space locations. $\textbf{y}^{\text{real}}$ and $\textbf{y}^{\text{imag}}$ represent the real and imaginary part, respectively. Accordingly, given the multi-coil k-space format of two outputs, $\textbf{y}_{\text{rec}\_\text{f}}$ and $\textbf{y}_{\text{rec}\_\text{p}}$, the overall loss function is formulated as:
\begin{equation}
\begin{aligned}
\mathcal{L} =\; & \lambda_{1}\mathcal{L}_{\text{NIK}}(\textbf{y}_{\text{rec}\_\text{f}}, \textbf{y})
+ \lambda_{2}\mathcal{L}_{\text{NIK}}(\textbf{y}_{\text{rec}\_\text{p}}, \textbf{y}) \\
& + \lambda_{3}\mathcal{L}_{\text{RelL2}}(\textbf{y}_{\text{rec}\_\text{f}}, \textbf{y})
+ \lambda_{4}\mathcal{L}_{\text{RelL2}}(\textbf{y}_{\text{rec}\_\text{p}}, \textbf{y}) .
\end{aligned}
\end{equation}
where $\lambda$s are hyper-parameters that control the relative weight of each loss term. In our experiments, we set $\lambda_{1} = \lambda_{2} = \lambda_{3} = \lambda_{4} = 0.5$ to treat all terms as equally important.

\section{Experimental Setup}
\subsection{Datasets}
To comprehensively assess the performance of the proposed I-FP-INR method, we evaluated them on two datasets: a public CMRxRecon2024 data\footnote{Available from the \href{https://www.synapse.org/Synapse:syn54951257/datasets/}{Synapse repository}.}~\cite{29} and an in-house dataset, which has been approved for research purposes by the institutional review board.

\begin{figure}[tb!]
\centering
\resizebox{0.46\textwidth}{!}{\includegraphics{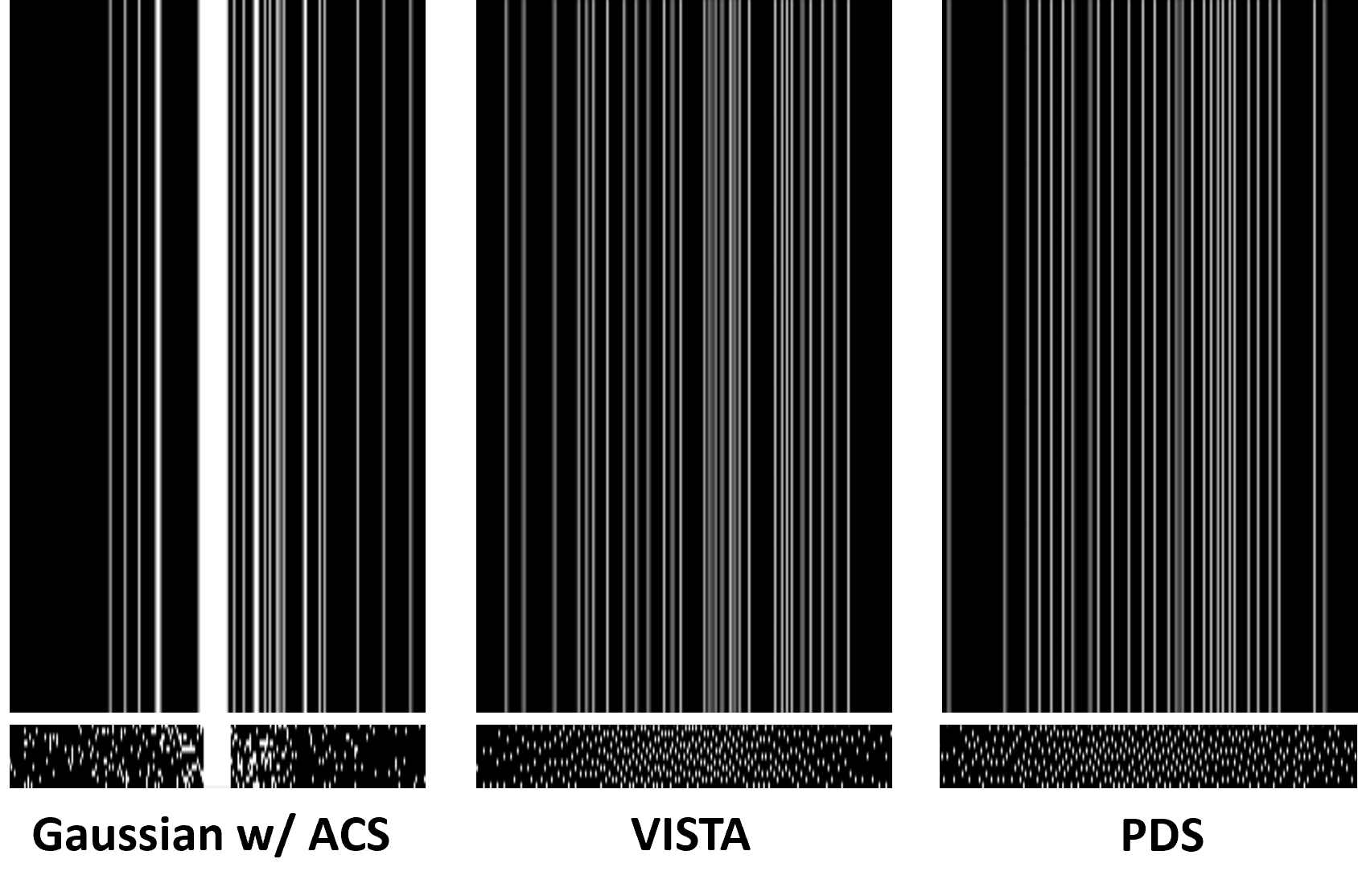}}
\caption{Visualization examples of three sampling mask types at R=8. The bottom panel shows sampling along the temporal dimension.} \label{fig_mask}
\end{figure}

For the public CMRxRecon2024 dataset, acquisitions were performed on a 3T MAGNETOM Vida scanner (Siemens Healthineers) using multi-channel cardiac coils. The dataset includes 330 healthy subjects, covering multiple contrasts, anatomical views, and k-space undersampling trajectories. As this work focuses on cardiac cine MRI, we use long-axis (LAX) and short-axis (SAX) views acquired with a balanced SSFP (bSSFP) sequence. Each subject includes 3 LAX slices and 8–14 SAX slices, each with 30 coils. For LAX acquisitions, the main parameters are: $\text{FOV}_{\text{x}}$ = 340–383 mm, $\text{FOV}_{\text{y}}$ = 236.79–379.58 mm, slice thickness = 6 mm, TR = 39.96–43.80 ms, TE = 1.46–1.57 ms, and flip angle = 39–52°. For SAX acquisitions, the parameters are: $\text{FOV}_{\text{x}}$ = 344–404 mm, $\text{FOV}_{\text{y}}$ = 215–404 mm, slice thickness = 8 mm, TR = 45.78–47.88 ms, TE = 1.44–1.50 ms, and flip angle = 37–44°. Since the INR method operates at the subject level, 16 subjects were randomly selected for evaluation. The dataset provides three temporally variable sampling trajectories: uniform Cartesian, Gaussian Cartesian, and pseudo-radial, with acceleration factors ranging from 4 to 24. As variable-density (VD) pattern is more practical, we only used the provided Gaussian Cartesian sampling pattern with an Auto-Calibration Signal (ACS) region of 16 lines, which corresponds to VD random sampling with a fully sampled k-space center. We also simulated two additional Cartesian sampling patterns without a fully-sampled k-space center: a variable density incoherent spatiotemporal acquisition (VISTA) pattern~\cite{40}, which optimizes k–t sampling with increased sampling density toward the center of k-space, and a Poisson disk sampling (PDS) pattern~\cite{39}, implemented along the phase-encoding direction with a minimum-distance constraint, resulting in a more uniform sample distribution while adhering to a variable-density profile. Fig.~\ref{fig_mask} shows these three kinds of sampling masks to give an overview.
Since an acceleration factor of 8 is already challenging in the actual application, we primarily evaluated our method at acceleration factors of 4 and 8 to reflect realistic, practically relevant scenarios.



For the in-house LUMC data, scans were acquired on a prototype 0.6T BlueSeal Philips scanner (Best, the Netherlands) equipped with 45mT/m gradients (slew max 200 mT/m/ms) and body receive arrays (posterior and anterior, 16 elements each). Subjects were scanned in short-axis (SAX) views using a balanced SSFP (bSSFP) sequence and a coil array with 27 coil elements. The main acquisition parameters are: FOV = 300 $\times$ 300 mm$^{2}$, voxel size = 1.25 $\times$ 1.25  mm$^{2}$, TR = 4.0 ms, TE = 2.0 ms, slice thickness = 8 mm, and flip angle = 90°. Variable-density sampling masks based on a Poisson disk distribution were applied to emphasize low-frequency k-space region, with acceleration factors ranging from 3 to 6. Breath-hold acquisition was performed during scanning to reduce respiratory motion artifacts. Notably, a fixed sampling mask was applied across all temporal frames, making reconstruction more challenging.

\subsection{Implementation Details}
Before optimization with the I-FP-INR method, several pre-processing steps are applied. The input images were normalized by their maximum absolute value, and coordinates were scaled to $[-1, 1]$. Due to the large size of the public data ($\sim$500 $\times$ 250), using all coordinates for the entire image sequence would impose a substantial computational burden. To address this, we adopted a gradient accumulation strategy: each optimization iteration is split across frames, processing one frame’s coordinates at a time while using the full cine sequence to compute loss and gradients. This highly reduces GPU memory usage and improves convergence efficiency.

The models were implemented in PyTorch 2.0.0 and trained on an NVIDIA A100 GPU with 40GB of memory. Hyper-parameters were kept unchanged for the two datasets. I-FP-INR was optimized using the AdamW optimizer with parameters $\beta_{1}=0.9$, $\beta_{2}=0.999$, $\epsilon=10^{-8}$. The initial learning rate was set to $1 \times 10^{-4}$ with weight decay 0.1, decayed by a factor of 0.95 every 50 iterations. Inference stage was applied every 50 iterations, with 300 total iterations to ensure full convergence. As mentioned earlier, the positional encoding for each coordinate was divided into a spatial component and a temporal component. Following~\cite{25}, we assigned a larger dimension of 480 to the spatial component, while the temporal component has a dimension of 96, yielding a total encoding length of 576 after concatenation. Each branch employs a 7-layer MLP. Then for comprehensively evaluating the public dataset, we used Peak Signal-to-Noise Ratio (PSNR), Structural Similarity Index (SSIM), Deep Image Structure and Texture Similarity (DISTS)~\cite{35}, and the Haar Wavelet-based Perceptual Similarity Index (HaarPSI)~\cite{36}. While PSNR and SSIM measure reconstruction error and structural similarity relative to a reference, DISTS and HaarPSI can better capture perceptual similarity aligned with radiological assessments~\cite{37}, reflecting potential clinical relevance. Statistical significance was assessed using the Wilcoxon signed-rank test.

\section{Results}
\subsection{Quantitative and Qualitative Comparison with Baselines}
To comprehensively evaluate I-FP-INR, we compared it with one traditional reconstruction method as well as several related approaches using INR priors: L+S~\cite{38}, P-INR~\cite{27}, KP-INR~\cite{27}, Hash-INR~\cite{24}, and I-P-INR. Specifically: (1) L+S is a well known traditional dynamic MRI reconstruction method that decomposes data into low-rank and sparse components; (2) P-INR, an INR-based k-space method that follows the NIK~\cite{14} framework by learning the mapping from coordinates to sampled k-space values; (3) KP-INR, a predecessor of I-FP-INR that introduces a k-space feature branch interacting with the original coordinate branch to guide the optimization; (4) Hash-INR is a representative image-domain INR-based method for dynamic MRI reconstruction, using hash-grid encoding to map coordinates to the embedding space and showing good image quality at high accelerations; (5) I-P-INR is a single-branch version of I-FP-INR without the feature branch, directly mapping coordinates to the reconstructed coil-combined image using the same loss of I-FP-INR. It also incorporates the CG algorithm during final inference for data consistency, enabling assessment of the feature branch’s impact. Furthermore, I-FP-INR was evaluated using two reconstruction strategies to systematically compare them under different conditions.

Table~\ref{tab1} summarizes the quantitative performance of I-FP-INR on the public dataset. Overall, the multi-coil strategy performed best under the Gaussian Cartesian sampling pattern with a fully-sampled k-space center. In the other two settings, the coil-combined strategy performed better with higher metric values. Although KP-INR achieved competitive metrics under VISTA sampling at R=4, visual inspection shows that the reconstructed images still contain some noticeable artifacts. Furthermore, despite the strong performance of I-P-INR across various situations, I-FP-INR consistently achieves further improvements, highlighting the effectiveness of incorporating image feature embeddings.

\begin{table*}[tb!]
\caption{Quantitative results of the proposed I-FP-INR compared with baseline models on the CMRxRecon2024 dataset. Here, "VD w/ ACS" points to the provided Gaussian Cartesian pattern from public data. "MC" and "CC" denote the multi-coil and coil-combined reconstruction strategy, respectively. The best results are highlighted in \textbf{bold}. $\dagger$ denotes statistical significance ($p < 0.0083$, paired Wilcoxon signed-rank test with Bonferroni correction) compared to I-FP-INR (CC).
\label{tab1}}
\centering
\resizebox{0.8\textwidth}{!}{
\begin{tabular}{c|c|l|c|c|c|c}
\hline
R & Sampling Pattern & Recon Method & PSNR $\uparrow$ & SSIM $\uparrow$ & DISTS $\uparrow$ & HaarPSI $\uparrow$ \\
\hline
\multirow{21}{*}{4$\times$} & \multirow{7}{*}{\centering{VD w/ ACS}} & L+S & $41.25 \pm 2.09\textsuperscript{$\dagger$}$ & $0.969 \pm 0.011\textsuperscript{$\dagger$}$ & $0.945 \pm 0.015\textsuperscript{$\dagger$}$ & $0.934 \pm 0.033\textsuperscript{$\dagger$}$ \\
\multirow{7}{*}{} & \multirow{7}{*}{} & P-INR & $41.01 \pm 1.98$\textsuperscript{$\dagger$} & $0.971 \pm 0.009$\textsuperscript{$\dagger$} & $0.953 \pm 0.014$\textsuperscript{$\dagger$} & $0.952 \pm 0.025$ \\
\multirow{7}{*}{} & \multirow{7}{*}{} & KP-INR & $42.97 \pm 2.08$ & $0.981 \pm 0.006$\textsuperscript{$\dagger$} & $0.947 \pm 0.011$\textsuperscript{$\dagger$} & $0.965 \pm 0.016$\textsuperscript{$\dagger$} \\
\multirow{7}{*}{} & \multirow{7}{*}{} & Hash-INR & $35.04 \pm 3.30$\textsuperscript{$\dagger$} & $0.894 \pm 0.046$\textsuperscript{$\dagger$} & $0.883 \pm 0.033$\textsuperscript{$\dagger$} & $0.813 \pm 0.091$\textsuperscript{$\dagger$} \\
\multirow{7}{*}{} & \multirow{7}{*}{} & I-P-INR & $42.06 \pm 2.00$\textsuperscript{$\dagger$} & $0.976 \pm 0.008$\textsuperscript{$\dagger$} & $0.953 \pm 0.013$\textsuperscript{$\dagger$} & $0.948 \pm 0.023$\textsuperscript{$\dagger$} \\
\multirow{7}{*}{} & \multirow{7}{*}{} & I-FP-INR (MC) & $\mathbf{45.11} \pm \mathbf{2.20}$\textsuperscript{$\dagger$} & $\mathbf{0.983} \pm \mathbf{0.006}$\textsuperscript{$\dagger$} & $\mathbf{0.965} \pm \mathbf{0.009}$\textsuperscript{$\dagger$} & $\mathbf{0.977} \pm \mathbf{0.011}$\textsuperscript{$\dagger$} \\
\multirow{7}{*}{} & \multirow{7}{*}{} & I-FP-INR (CC) & $42.91 \pm 2.13$ & $0.978 \pm 0.008$ & $0.955 \pm 0.013$ & $0.951 \pm 0.023$ \\
\cline{2-7}

\multirow{7}{*}{} & \multirow{7}{*}{VISTA} & L+S & $40.23 \pm 1.94\textsuperscript{$\dagger$}$ & $0.968 \pm 0.011\textsuperscript{$\dagger$}$ & $0.947 \pm 0.015\textsuperscript{$\dagger$}$ & $0.923 \pm 0.029\textsuperscript{$\dagger$}$ \\
\multirow{7}{*}{} & \multirow{7}{*}{} & P-INR & $38.70 \pm 2.69$\textsuperscript{$\dagger$} & $0.950 \pm 0.020$\textsuperscript{$\dagger$} & $0.934 \pm 0.015$\textsuperscript{$\dagger$} & $0.918 \pm 0.029$\textsuperscript{$\dagger$} \\
\multirow{7}{*}{} & \multirow{7}{*}{} & KP-INR & $40.99 \pm 2.32$ & $\mathbf{0.971} \pm \mathbf{0.010}$ & $\mathbf{0.953} \pm \mathbf{0.011}$ & $\mathbf{0.950} \pm \mathbf{0.019}$\textsuperscript{$\dagger$} \\
\multirow{7}{*}{} & \multirow{7}{*}{} & Hash-INR & $36.20 \pm 3.46$\textsuperscript{$\dagger$} & $0.889 \pm 0.052$\textsuperscript{$\dagger$} & $0.895 \pm 0.037$\textsuperscript{$\dagger$} & $0.836 \pm 0.099$\textsuperscript{$\dagger$} \\
\multirow{7}{*}{} & \multirow{7}{*}{} & I-P-INR & $40.08 \pm 2.16$\textsuperscript{$\dagger$} & $0.965 \pm 0.013$\textsuperscript{$\dagger$} & $0.948 \pm 0.015$\textsuperscript{$\dagger$} & $0.929 \pm 0.031$\textsuperscript{$\dagger$} \\
\multirow{7}{*}{} & \multirow{7}{*}{} & I-FP-INR (MC) & $33.25 \pm 2.56$\textsuperscript{$\dagger$} & $0.828 \pm 0.041$\textsuperscript{$\dagger$} & $0.896 \pm 0.018$\textsuperscript{$\dagger$} & $0.790 \pm 0.056$\textsuperscript{$\dagger$} \\
\multirow{7}{*}{} & \multirow{7}{*}{} & I-FP-INR (CC) & $\mathbf{41.07} \pm \mathbf{2.53}$ & $0.970 \pm 0.013$ & $0.952 \pm 0.015 $ & $0.939 \pm 0.033$ \\
\cline{2-7}
\multirow{7}{*}{} & \multirow{7}{*}{PDS} & L+S & $34.19 \pm 2.30\textsuperscript{$\dagger$}$ & $0.890 \pm 0.039\textsuperscript{$\dagger$}$ & $0.877 \pm 0.027\textsuperscript{$\dagger$}$ & $0.758 \pm 0.063\textsuperscript{$\dagger$}$ \\
\multirow{7}{*}{} & \multirow{7}{*}{} & P-INR & $33.92 \pm 3.18$\textsuperscript{$\dagger$} & $0.876 \pm 0.044$\textsuperscript{$\dagger$} & $0.871 \pm 0.028$\textsuperscript{$\dagger$} & $0.782 \pm 0.066$\textsuperscript{$\dagger$} \\
\multirow{7}{*}{} & \multirow{7}{*}{} & KP-INR & $35.07 \pm 3.00$\textsuperscript{$\dagger$} & $0.912 \pm 0.037$\textsuperscript{$\dagger$} & $0.896 \pm 0.026$\textsuperscript{$\dagger$} & $0.820 \pm 0.063$\textsuperscript{$\dagger$} \\
\multirow{7}{*}{} & \multirow{7}{*}{} & Hash-INR & $35.22 \pm 2.32$\textsuperscript{$\dagger$} & $0.887 \pm 0.051$\textsuperscript{$\dagger$} & $0.892 \pm 0.037$\textsuperscript{$\dagger$} & $0.811 \pm 0.096$\textsuperscript{$\dagger$} \\
\multirow{7}{*}{} & \multirow{7}{*}{} & I-P-INR & $38.07 \pm 2.23$\textsuperscript{$\dagger$} & $0.955 \pm 0.018$\textsuperscript{$\dagger$} & $0.936 \pm 0.018$\textsuperscript{$\dagger$} & $0.892 \pm 0.041$ \\
\multirow{7}{*}{} & \multirow{7}{*}{} & I-FP-INR (MC) & $30.60 \pm 2.78$\textsuperscript{$\dagger$} & $0.782 \pm 0.071$\textsuperscript{$\dagger$} & $0.875 \pm 0.028$\textsuperscript{$\dagger$} & $0.733 \pm 0.077$\textsuperscript{$\dagger$} \\
\multirow{7}{*}{} & \multirow{7}{*}{} & I-FP-INR (CC) & $\mathbf{39.02} \pm \mathbf{2.57}$ & $\mathbf{0.960} \pm \mathbf{0.019}$ & $\mathbf{0.941} \pm \mathbf{0.019}$ & $\mathbf{0.904} \pm \mathbf{0.046}$ \\
\hline
\hline

\multirow{21}{*}{$8 \times$} & \multirow{7}{*}{VD w/ ACS} & L+S & $37.98 \pm 2.06$\textsuperscript{$\dagger$ } & $0.940 \pm 0.018$\textsuperscript{$\dagger$ } & $0.909 \pm 0.018$\textsuperscript{$\dagger$ } & $0.868 \pm 0.048$\textsuperscript{$\dagger$ } \\
\multirow{7}{*}{} & \multirow{7}{*}{} & P-INR & $35.78 \pm 2.76$\textsuperscript{$\dagger$ } & $0.918 \pm 0.025$\textsuperscript{$\dagger$ } & $0.877 \pm 0.024$\textsuperscript{$\dagger$ } & $0.862 \pm 0.046$\textsuperscript{$\dagger$ } \\
\multirow{7}{*}{} & \multirow{7}{*}{} & KP-INR & $38.98 \pm 2.08$\textsuperscript{$\dagger$ } & $0.956 \pm 0.014$\textsuperscript{$\dagger$ } & $0.911 \pm 0.016$\textsuperscript{$\dagger$ } & $0.908 \pm 0.032$\textsuperscript{$\dagger$ } \\
\multirow{7}{*}{} & \multirow{7}{*}{} & Hash-INR & $31.56 \pm 3.35$\textsuperscript{$\dagger$ } & $0.843 \pm 0.057$\textsuperscript{$\dagger$ } & $0.848 \pm 0.034$\textsuperscript{$\dagger$ } & $0.699 \pm 0.107$\textsuperscript{$\dagger$ } \\
\multirow{7}{*}{} & \multirow{7}{*}{} & I-P-INR & $38.70 \pm 2.13$\textsuperscript{$\dagger$ } & $0.957 \pm 0.015$\textsuperscript{$\dagger$ } & $0.923 \pm 0.018$\textsuperscript{$\dagger$ } & $0.907 \pm 0.043$\textsuperscript{$\dagger$ } \\
\multirow{7}{*}{} & \multirow{7}{*}{} & I-FP-INR (MC) & $\mathbf{42.01} \pm \mathbf{2.02}$\textsuperscript{$\dagger$ } & $\mathbf{0.973} \pm \mathbf{0.009}$\textsuperscript{$\dagger$ } & $\mathbf{0.943} \pm \mathbf{0.013}$\textsuperscript{$\dagger$ } & $\mathbf{0.949} \pm \mathbf{0.022}$\textsuperscript{$\dagger$ } \\
\multirow{7}{*}{} & \multirow{7}{*}{} & I-FP-INR (CC) & $40.03 \pm 2.13$ & $0.964 \pm 0.013$ & $0.934 \pm 0.017$ & $0.914 \pm 0.041$ \\
\cline{2-7}
\multirow{7}{*}{} & \multirow{7}{*}{VISTA} & L+S & $33.62 \pm 2.02$\textsuperscript{$\dagger$ } & $0.879 \pm 0.031$\textsuperscript{$\dagger$ } & $0.871 \pm 0.022$\textsuperscript{$\dagger$ } & $0.756 \pm 0.056$\textsuperscript{$\dagger$ } \\
\multirow{7}{*}{} & \multirow{7}{*}{} & P-INR & $32.62 \pm 3.60$\textsuperscript{$\dagger$ } & $0.855 \pm 0.045$\textsuperscript{$\dagger$ } & $0.856 \pm 0.033$\textsuperscript{$\dagger$ } & $0.774 \pm 0.076$\textsuperscript{$\dagger$ } \\
\multirow{7}{*}{} & \multirow{7}{*}{} & KP-INR & $35.15 \pm 2.73$\textsuperscript{$\dagger$ } & $0.915 \pm 0.028$\textsuperscript{$\dagger$ } & $0.894 \pm 0.025$\textsuperscript{$\dagger$ } & $0.838 \pm 0.054$\textsuperscript{$\dagger$ } \\
\multirow{7}{*}{} & \multirow{7}{*}{} & Hash-INR & $32.45 \pm 3.24$\textsuperscript{$\dagger$ } & $0.836 \pm 0.064$\textsuperscript{$\dagger$ } & $0.865 \pm 0.039$\textsuperscript{$\dagger$ } & $0.732 \pm 0.112$\textsuperscript{$\dagger$ } \\
\multirow{7}{*}{} & \multirow{7}{*}{} & I-P-INR & $37.25 \pm 2.20$\textsuperscript{$\dagger$ } & $0.945 \pm 0.021$\textsuperscript{$\dagger$ } & $0.928 \pm 0.019$\textsuperscript{$\dagger$ } & $0.884 \pm 0.045$\textsuperscript{$\dagger$ } \\
\multirow{7}{*}{} & \multirow{7}{*}{} & I-FP-INR (MC) & $31.30 \pm 3.20$\textsuperscript{$\dagger$ } & $0.793 \pm 0.068$\textsuperscript{$\dagger$} & $0.881 \pm 0.029$\textsuperscript{$\dagger$ } & $0.755 \pm 0.088$\textsuperscript{$\dagger$ } \\
\multirow{7}{*}{} & \multirow{7}{*}{} & I-FP-INR (CC) & $\mathbf{38.30} \pm \mathbf{2.61}$ & $\mathbf{0.953} \pm \mathbf{0.021}$ & $\mathbf{0.934} \pm \mathbf{0.019}$ & $\mathbf{0.896} \pm \mathbf{0.051}$ \\
\cline{2-7}
\multirow{7}{*}{} & \multirow{7}{*}{PDS} & L+S & $32.20 \pm 2.14$\textsuperscript{$\dagger$ } & $0.852 \pm 0.038$\textsuperscript{$\dagger$ } & $0.853 \pm 0.026$\textsuperscript{$\dagger$ } & $0.708 \pm 0.063$\textsuperscript{$\dagger$ } \\
\multirow{7}{*}{} & \multirow{7}{*}{} & P-INR & $31.33 \pm 3.63$\textsuperscript{$\dagger$ } & $0.822 \pm 0.062$\textsuperscript{$\dagger$ } & $0.835 \pm 0.047$\textsuperscript{$\dagger$ } & $0.717 \pm 0.010$\textsuperscript{$\dagger$ } \\
\multirow{7}{*}{} & \multirow{7}{*}{} & KP-INR & $33.68 \pm 3.12$\textsuperscript{$\dagger$ } & $0.893 \pm 0.038$\textsuperscript{$\dagger$ } & $0.882 \pm 0.028$\textsuperscript{$\dagger$ } & $0.796 \pm 0.063$\textsuperscript{$\dagger$ } \\
\multirow{7}{*}{} & \multirow{7}{*}{} & Hash-INR & $33.45 \pm 2.89$\textsuperscript{$\dagger$ } & $0.857 \pm 0.055$\textsuperscript{$\dagger$ } & $0.878 \pm 0.035$\textsuperscript{$\dagger$ } & $0.773 \pm 0.098$\textsuperscript{$\dagger$ } \\
\multirow{7}{*}{} & \multirow{7}{*}{} & I-P-INR & $36.90 \pm 2.32$\textsuperscript{$\dagger$ } & $0.943 \pm 0.021$\textsuperscript{$\dagger$ } & $0.926 \pm 0.020$\textsuperscript{$\dagger$ } & $0.874 \pm 0.047$\textsuperscript{$\dagger$ } \\
\multirow{7}{*}{} & \multirow{7}{*}{} & I-FP-INR (MC) & $28.78 \pm 2.84$\textsuperscript{$\dagger$ } & $0.752 \pm 0.069$\textsuperscript{$\dagger$ } & $0.855 \pm 0.029$\textsuperscript{$\dagger$ } & $0.672 \pm 0.076$\textsuperscript{$\dagger$ } \\
\multirow{7}{*}{} & \multirow{7}{*}{} & I-FP-INR (CC) & $\mathbf{37.94} \pm \mathbf{2.62} $ & $\mathbf{0.951}\pm \mathbf{0.023}$ & $\mathbf{0.931} \pm \mathbf{0.020}$ & $\mathbf{0.885} \pm \mathbf{0.054}$ \\
\cline{2-7}
\hline
\end{tabular}
}
\end{table*}

\begin{figure*}[tb!]
\centering
\resizebox{0.88\textwidth}{!}{\includegraphics{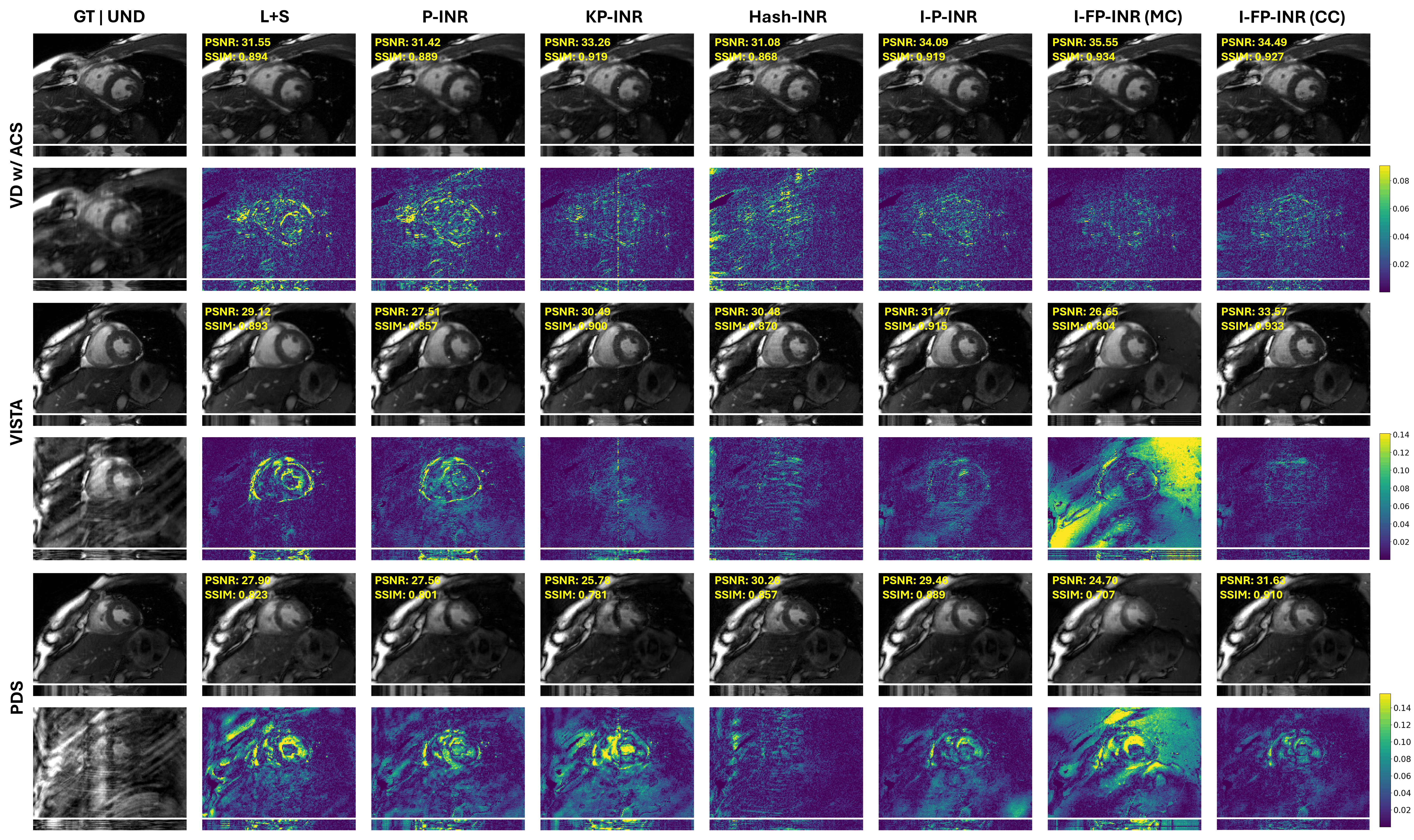}}
\caption{Qualitative comparison of the proposed I-FP-INR with baseline methods at R=4 under three sampling patterns. “GT” denotes ground truth and “UND” the undersampled input. The bottom panels show temporal profiles along the central line, while the top-left inset reports metrics for the cropped region. } \label{fig_r4}
\end{figure*}

\begin{figure*}[tb!]
\centering
\resizebox{0.88\textwidth}{!}{\includegraphics{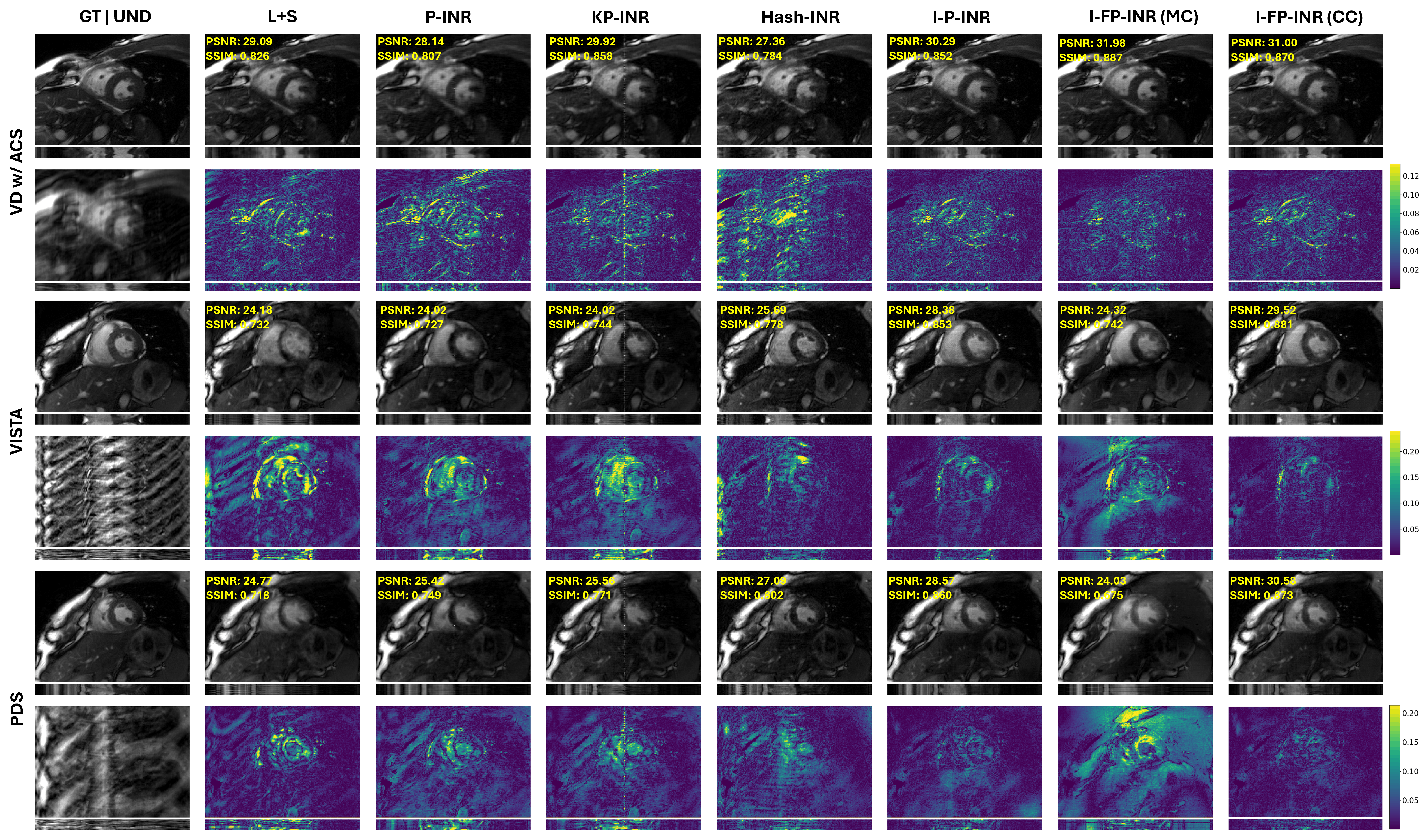}}
\caption{Qualitative comparison of the proposed I-FP-INR with baseline methods at R=8 under three sampling patterns.} \label{fig_r8}
\end{figure*}

Fig.~\ref{fig_r4} and \ref{fig_r8} present qualitative visualizations across all methods, with cine sequences cropped to highlight the cardiac region. Consistent with Table~\ref{tab1}, I-FP-INR achieved the highest metrics values computed within the cropped regions. The multi-coil strategy attained higher metrics and better image qualities under Gaussian Cartesian sampling with a fully-sampled center region, but struggled to accurately estimate pixel values in other patterns, leading to contrast discrepancies despite effective artifact removal. In contrast, the coil-combined strategy produced comparable visual quality under the Gaussian pattern while demonstrating greater robustness across VISTA and PDS patterns. At R=4, L+S yielded reconstructions without severe artifacts, but fine cardiac structures remained noticeably blurred, as also reflected in the temporal section. Under more challenging sampling (e.g., PDS), the method failed to fully suppress residual artifacts, indicating limited reconstruction capability. These limitations worsened at R=8, with substantially increased artifacts and degraded image fidelity. For k-space INR methods, incorporating feature embeddings improved reconstruction sharpness and enhanced fine detail recovery. Nevertheless, systematic artifacts persisted, including a bright central spot or axial streaking, indicating inconsistencies in modeling low-frequency k-space components. Although Hash-INR demonstrated competitive metrics at R=4 and R=8, residual artifacts remained visible. Similarly, I-P-INR attained performance comparable to I-FP-INR in both metrics and visualization, but remained less effective in recovering fine structural details.

We also recorded the average reconstruction time per slice for each method. The traditional L+S approach is the fastest, requiring less than 1 minute. Among the INR-based methods, the image-domain approaches based on the original design, Hash-INR and I-P-INR, are relatively efficient, taking approximately 14 and 12 minutes, respectively. Although Hash-INR has a lighter design, it requires more iterations in its original setting, leading to longer run times. The two I-FP-INR variants exhibit similar reconstruction times: the coil-combined strategy requires about 44 minutes and the multi-coil variant about 45 minutes. In contrast, the two k-space domain INR methods are the slowest, as they require much more optimization iterations to converge, resulting in reconstruction times of roughly 101 minutes for P-INR and 305 minutes for KP-INR.

\subsection{Ablation Studies}
In this work, I-FP-INR augments the original coordinate branch with an additional feature branch, enabling flexible activation functions and positional encoding. A combined loss enforces data consistency in the measured k-space, and a two-stage alternating optimization strategy progressively refines feature embeddings during optimization. To evaluate each component, we conducted a comprehensive ablation studies on the I-FP-INR with coil-combined reconstruction strategy under VISTA sampling at two acceleration factors.

\begin{table*}[!tbh]
\centering
\caption{Quantitative results of the ablation study on different design choices for each INR branch of I-FP-INR (CC) on the CMRxRecon2024 dataset. Best results are highlighted in \textbf{bold}. $\dagger$ denotes statistical significance ($p < 0.0125$, paired Wilcoxon signed-rank test with Bonferroni correction) compared to I-FP-INR (CC), which employs WIRE in the I-P-INR branch and ReLU MLP in the I-F-INR branch.}
\label{tab2_ablation1}
\resizebox{1\textwidth}{!}{
\begin{tabular}{c c c c c c c c c c}
\toprule
\multirow{2}{*}{I-P-INR} & \multirow{2}{*}{I-F-INR} & \multicolumn{4}{c}{$4 \times$} & \multicolumn{4}{c}{$8 \times$} \\
\cmidrule(lr){3-6} \cmidrule(lr){7-10}
 &  & PSNR$\uparrow$ & SSIM$\uparrow$ & DISTS$\uparrow$ & HaarPSI$\uparrow$ & PSNR$\uparrow$ & SSIM$\uparrow$ & DISTS$\uparrow$ &  HaarPSI$\uparrow$ \\
\hline
ReLU MLP & ReLU MLP & $33.99 \pm 2.17$\textsuperscript{$\dagger$ } & $0.883 \pm 0.034$\textsuperscript{$\dagger$ }  &$0.890 \pm 0.018$\textsuperscript{$\dagger$ } & $0.836 \pm 0.037$\textsuperscript{$\dagger$ } & $31.15 \pm 2.26$\textsuperscript{$\dagger$ } & $0.834 \pm 0.041$\textsuperscript{$\dagger$ } & $0.855 \pm 0.020$\textsuperscript{$\dagger$ } & $0.736 \pm 0.050$\textsuperscript{$\dagger$ } \\
SIREN & ReLU MLP & $35.17 \pm 2.02$\textsuperscript{$\dagger$ } & $0.914 \pm 0.025$\textsuperscript{$\dagger$ } & $0.904 \pm 0.015$\textsuperscript{$\dagger$ } & $0.874 \pm 0.028$\textsuperscript{$\dagger$ } & $32.23 \pm 2.04$\textsuperscript{$\dagger$ } & $0.865 \pm 0.034$\textsuperscript{$\dagger$ } & $0.859 \pm 0.017$\textsuperscript{$\dagger$ } & $0.793 \pm 0.037$\textsuperscript{$\dagger$ } \\
\hline
WIRE & SIREN & $39.97 \pm 2.40$\textsuperscript{$\dagger$ } & $0.963 \pm 0.024$\textsuperscript{$\dagger$ } & $0.945 \pm 0.021$\textsuperscript{$\dagger$ } & $0.925 \pm 0.043$\textsuperscript{$\dagger$ } & $37.25 \pm 2.22$\textsuperscript{$\dagger$ } & $0.946 \pm 0.021$\textsuperscript{$\dagger$ } & $0.930 \pm 0.018$\textsuperscript{$\dagger$ } & $0.883 \pm 0.045$\textsuperscript{$\dagger$ } \\
WIRE & WIRE & $40.54 \pm 2.23$\textsuperscript{$\dagger$ } & $0.969 \pm 0.012$\textsuperscript{$\dagger$ } & $0.949 \pm 0.014$\textsuperscript{$\dagger$ } & $0.935 \pm 0.030$\textsuperscript{$\dagger$ } & $37.74 \pm 2.36$\textsuperscript{$\dagger$ } & $0.952 \pm 0.020$ & $0.933 \pm 0.019$\textsuperscript{$\dagger$ } & $0.891 \pm 0.048$\textsuperscript{$\dagger$ }\\
\hline
WIRE & ReLU MLP & $\mathbf{41.07} \pm \mathbf{2.53}$ & $\mathbf{0.970} \pm \mathbf{0.013}$ & $\mathbf{0.952} \pm \mathbf{0.015}$ & $\mathbf{0.939} \pm \mathbf{0.033}$ & $\mathbf{38.30} \pm \mathbf{2.61}$ & $\mathbf{0.953} \pm \mathbf{0.021}$ & $\mathbf{0.934} \pm \mathbf{0.019}$ & $\mathbf{0.896} \pm \mathbf{0.051}$\\
\bottomrule
\end{tabular}
}
\end{table*}
Table~\ref{tab2_ablation1} summarizes ablation results for different INR choices in the two branches. We evaluated ReLU MLP~\cite{32}, SIREN~\cite{41}, and WIRE~\cite{34}, by varying one branch while fixing the other. Overall, WIRE in the coordinate branch and ReLU MLP in the feature branch achieved the best performance across both acceleration factors. This can be attributed to their complementary properties: WIRE effectively captures multi-scale signals with both high- and low-frequency components, while ReLU MLP efficiently learns feature embeddings. Using WIRE in both branches also performed well, confirming its advantage over SIREN and ReLU MLP in modeling complex, multi-scale signals. Notably, the I-P-INR branch played a more critical role: changing its INR type lead to a substantial performance drop, whereas similar modifications in the I-F-INR branch had a smaller impact. This is expected, as the I-P-INR branch serves as the primary coordinate-to-signal mapping, while the I-F-INR branch is designed to provide complementary feature embeddings that further enrich the representation, rather than serve as the core mapping function. This observation is further supported by the comparisons between I-P-INR and I-FP-INR in Table~\ref{tab1} and qualitative results, highlighting the dominant role of the I-P-INR branch.

\begin{table*}[!tbh]
\centering
\caption{Quantitative results of the ablation study on different positional encoding combinations for P-INR branch of I-FP-INR (CC) on the CMRxRecon2024 dataset. "PE (S)" denotes the positional encoding applied to the spatial component, while “PE (T)” refers to application of the temporal component. Best results are highlighted in \textbf{bold}. $\dagger$ denotes statistical significance ($p < 0.0167$, paired Wilcoxon signed-rank test with Bonferroni correction) compared to I-FP-INR (CC), which employs NeRF-based positional encoding for the spatial component and Fourier encoding for the temporal component.}
\label{tab2_ablation2}
\resizebox{1\textwidth}{!}{
\begin{tabular}{c c c c c c c c c c}
\toprule
\multirow{2}{*}{PE (S)} & \multirow{2}{*}{PE (T)} & \multicolumn{4}{c}{$4 \times$} & \multicolumn{4}{c}{$8 \times$} \\
\cmidrule(lr){3-6} \cmidrule(lr){7-10}
&  & PSNR$\uparrow$ & SSIM$\uparrow$ & DISTS$\uparrow$ & HaarPSI$\uparrow$ & PSNR$\uparrow$ & SSIM$\uparrow$ & DISTS$\uparrow$ &  HaarPSI$\uparrow$ \\
\hline
Fourier & Fourier & $38.40 \pm 2.49$\textsuperscript{$\dagger$ } & $0.947 \pm 0.021$\textsuperscript{$\dagger$ } & $0.901 \pm 0.018$\textsuperscript{$\dagger$ } & $0.902 \pm 0.043$\textsuperscript{$\dagger$ } & $36.05 \pm 2.69$\textsuperscript{$\dagger$ } & $0.926 \pm 0.028$\textsuperscript{$\dagger$ } & $0.878 \pm 0.021$\textsuperscript{$\dagger$ } & $0.849 \pm 0.058$\textsuperscript{$\dagger$ } \\
Fourier & NeRF & $38.15 \pm 2.48$\textsuperscript{$\dagger$ } & $0.945 \pm 0.022$\textsuperscript{$\dagger$ } & $0.900 \pm 0.019$\textsuperscript{$\dagger$ } & $0.898 \pm 0.043$\textsuperscript{$\dagger$ } & $36.01 \pm 2.51$\textsuperscript{$\dagger$ } & $0.925 \pm 0.028$\textsuperscript{$\dagger$ } & $0.877 \pm 0.022$\textsuperscript{$\dagger$ } & $0.848 \pm 0.058$\textsuperscript{$\dagger$ } \\
NeRF & NeRF & $40.54 \pm 2.35$\textsuperscript{$\dagger$ } & $0.967 \pm 0.014$\textsuperscript{$\dagger$ } & $0.948 \pm 0.016$\textsuperscript{$\dagger$ } & $0.931 \pm 0.035$\textsuperscript{$\dagger$ } & $38.10 \pm 2.45$\textsuperscript{$\dagger$ } & $0.952 \pm 0.020$ & $0.934 \pm 0.019$ & $0.893 \pm 0.045$\textsuperscript{$\dagger$ } \\
\hline
NeRF & Fourier & $\mathbf{41.07} \pm \mathbf{2.53}$ & $\mathbf{0.970} \pm \mathbf{0.013}$ & $\mathbf{0.952} \pm \mathbf{0.015}$ & $\mathbf{0.939} \pm \mathbf{0.033}$ & $\mathbf{38.30} \pm \mathbf{2.61}$ & $\mathbf{0.953} \pm \mathbf{0.021}$ & $\mathbf{0.934} \pm \mathbf{0.019}$ & $\mathbf{0.896} \pm \mathbf{0.051}$\\
\bottomrule
\end{tabular}
}
\end{table*}
As described earlier, the coordinates are decomposed into spatial and temporal components with dedicated positional encoding. Table~\ref{tab2_ablation2} presents the ablation results for different positional encoding combinations, focusing on NeRF and Fourier encoding. Overall, the best performance is achieved with NeRF encoding for spatial coordinates and Fourier encoding for temporal coordinates. Although both use sinusoidal mappings, they differ in frequency coverage and density. Temporal dynamics are generally smooth and dominated by low-frequency variations, making them well-suited for Fourier encoding, whereas spatial signals contain richer high-frequency details that require the enhanced multi-scale representation provided by NeRF encoding. Furthermore, spatial encoding has a more pronounced impact on performance, as evidenced by the noticeable degradation when replacing NeRF encoding with standard Fourier encoding for the spatial component, further supporting the above explanation.
\begin{table*}[!tbh]
\centering
\caption{Quantitative results of the ablation study on the loss function design of I-FP-INR (CC) on the CMRxRecon2024 dataset. Best results are highlighted in \textbf{bold}. $\dagger$ denotes statistical significance ($p < 0.025$, paired Wilcoxon signed-rank test with Bonferroni correction) compared to I-FP-INR (CC), which applies the combination of NIK loss and Relative L2 loss.}
\label{tab2_ablation3}
\resizebox{1\textwidth}{!}{
\begin{tabular}{c c c c c c c c c}
\toprule
\multirow{2}{*}{Loss Function} &  \multicolumn{4}{c}{$4 \times$} & \multicolumn{4}{c}{$8 \times$} \\
\cmidrule(lr){2-5} \cmidrule(lr){6-9}
 & PSNR$\uparrow$ & SSIM$\uparrow$ & DISTS$\uparrow$ & HaarPSI$\uparrow$ & PSNR$\uparrow$ & SSIM$\uparrow$ & DISTS$\uparrow$ &  HaarPSI$\uparrow$ \\
\hline
NIK Loss & $40.78 \pm 2.78$\textsuperscript{$\dagger$ } & $0.968 \pm 0.015$\textsuperscript{$\dagger$ } & $0.951 \pm 0.016$\textsuperscript{$\dagger$ } & $0.938 \pm 0.028$ & $36.76 \pm 2.81$\textsuperscript{$\dagger$ } & $0.938 \pm 0.028$\textsuperscript{$\dagger$ } & $0.910 \pm 0.025$\textsuperscript{$\dagger$ } & $0.866 \pm 0.066$\textsuperscript{$\dagger$ } \\
RelL2 Loss & $39.83 \pm 2.30$\textsuperscript{$\dagger$ } & $0.964 \pm 0.014$\textsuperscript{$\dagger$ } & $0.943 \pm 0.015$\textsuperscript{$\dagger$ } & $0.920 \pm 0.034$\textsuperscript{$\dagger$ } & $37.43 \pm 2.37$\textsuperscript{$\dagger$ } & $0.947 \pm 0.023$\textsuperscript{$\dagger$ } & $0.930 \pm 0.019$\textsuperscript{$\dagger$ } & $0.876 \pm 0.047$\textsuperscript{$\dagger$ } \\
\hline
NIK Loss + RelL2 Loss & $\mathbf{41.07} \pm \mathbf{2.53}$ & $\mathbf{0.970} \pm \mathbf{0.013}$ & $\mathbf{0.952} \pm \mathbf{0.015}$ & $\mathbf{0.939} \pm \mathbf{0.033}$ & $\mathbf{38.30} \pm \mathbf{2.61}$ & $\mathbf{0.953} \pm \mathbf{0.021}$ & $\mathbf{0.934} \pm \mathbf{0.019}$ & $\mathbf{0.896} \pm \mathbf{0.051}$\\
\bottomrule
\end{tabular}
}
\end{table*}

Table~\ref{tab2_ablation3} examines the impact of loss functions on INR reconstruction. Under VISTA sampling, combining NIK loss and relative L2 loss yields the best performance. When used individually, NIK loss performs better at lower acceleration with dense low-frequency sampling, while relative L2 loss is more effective at higher acceleration. Overall, NIK loss emphasizes accurate magnitude estimation with adaptive weighting, whereas relative L2 loss explicitly enforces consistency in both magnitude and phase, leading to more balanced optimization. We hypothesize that dense low-frequency sampling contains phase information well, favoring magnitude-focused losses for enhancing contrast and edges. Higher acceleration makes k-space phase become noisier and more ambiguous, requiring phase-aware losses to correct.


\begin{table*}[!tbh]
\centering
\caption{Quantitative results of the ablation study on the impact of two-stage optimization design of I-FP-INR (CC) on the CMRxRecon2024 dataset. Best results are highlighted in \textbf{bold}. $\dagger$ denotes statistical significance ($p < 0.05$, paired Wilcoxon signed-rank test with Bonferroni correction) compared to I-FP-INR (CC), which contains two stages during the optimization.}
\label{tab2_ablation4}
\resizebox{1\textwidth}{!}{
\begin{tabular}{c c c c c c c c c}
\toprule
\multirow{2}{*}{Methods} &  \multicolumn{4}{c}{$4 \times$} & \multicolumn{4}{c}{$8 \times$} \\
\cmidrule(lr){2-5} \cmidrule(lr){6-9}
 & PSNR$\uparrow$ & SSIM$\uparrow$ & DISTS$\uparrow$ & HaarPSI$\uparrow$ & PSNR$\uparrow$ & SSIM$\uparrow$ & DISTS$\uparrow$ &  HaarPSI$\uparrow$ \\
\hline
I-FP-INR (CC) w/o Inference Stage & $40.50 \pm 2.32$\textsuperscript{$\dagger$ } & $0.967 \pm 0.014$\textsuperscript{$\dagger$ } & $0.948 \pm 0.016$\textsuperscript{$\dagger$ } & $0.932 \pm 0.034$\textsuperscript{$\dagger$ } & $37.95 \pm 2.33$\textsuperscript{$\dagger$ } & $0.951 \pm 0.019$\textsuperscript{$\dagger$ } & $0.933 \pm 0.018$ & $0.892 \pm 0.043$\textsuperscript{$\dagger$ } \\
I-FP-INR (CC) & $\mathbf{41.07} \pm \mathbf{2.53}$ & $\mathbf{0.970} \pm \mathbf{0.013}$ & $\mathbf{0.952} \pm \mathbf{0.015}$ & $\mathbf{0.939} \pm \mathbf{0.033}$ & $\mathbf{38.30} \pm \mathbf{2.61}$ & $\mathbf{0.953} \pm \mathbf{0.021}$ & $\mathbf{0.934} \pm \mathbf{0.019}$ & $\mathbf{0.896} \pm \mathbf{0.051}$\\
\bottomrule
\end{tabular}
}
\end{table*}
Finally, Table~\ref{tab2_ablation4} presents a comparison of models with and without incorporating the inference stage during optimization. Removing the inference stage leads to an overall performance degradation, although the extent of this drop decreases as the acceleration factor increases. This is likely because the branch with positional encoding plays a more dominant role, mitigating the impact of excluding the inference stage. However, the results still indicate the benefits of incorporating the inference stage, as it enables more accurate feature representations via continuous refinement of the input image.

\subsection{Qualitative Visualizations of LUMC In-house Data}
While prior performance comparisons and ablation studies provide comprehensive assessment and demonstrate the effectiveness of I-FP-INR, we further evaluated it on in-house data to validate real-world applicability and reinforce some previous observations. I-FP-INR was tested with two reconstruction strategies and compared against L+S. In the original acquisitions, a fixed variable-density sampling mask was applied across all frames, with acceleration factors ranging from 3 to 6. This restricts the INR method to a subset of k-space, with the remaining regions consistently unobserved across frames, thereby limiting inter-frame correlation modeling and the learning of the global cine k-space distribution. Therefore, we also simulated a frame-varying variable-density mask by splitting the fixed mask so that each frame retained half of the sampled lines. This setup enables exploration of higher acceleration and facilitates more effective utilization of spatio-temporal features in INR optimization.

\begin{figure*}[!tb]
\centering
\subfloat[]{\includegraphics[width=3.5in]{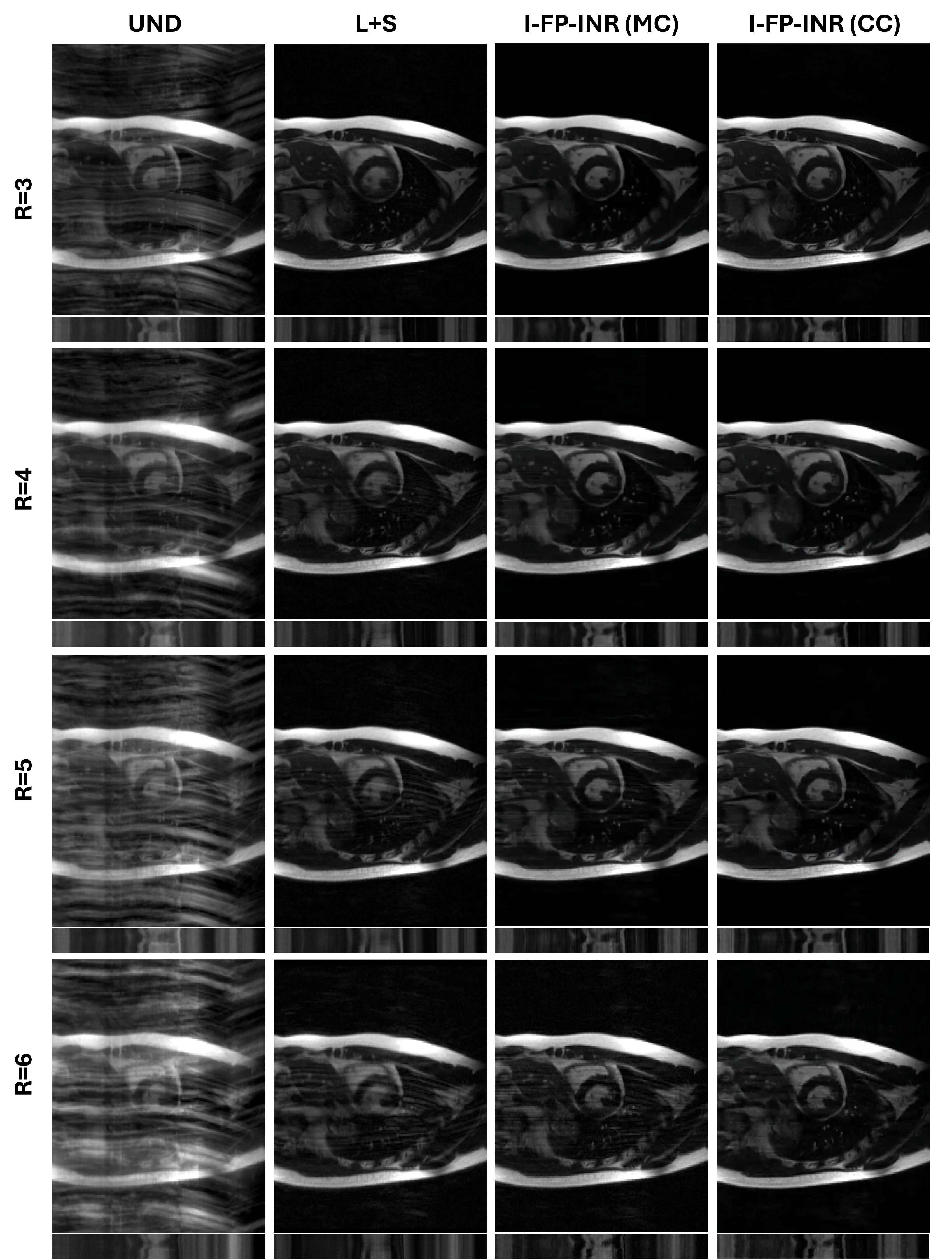}%
\label{fig_lumc_still}}
\hfil
\subfloat[]{\includegraphics[width=3.5in]{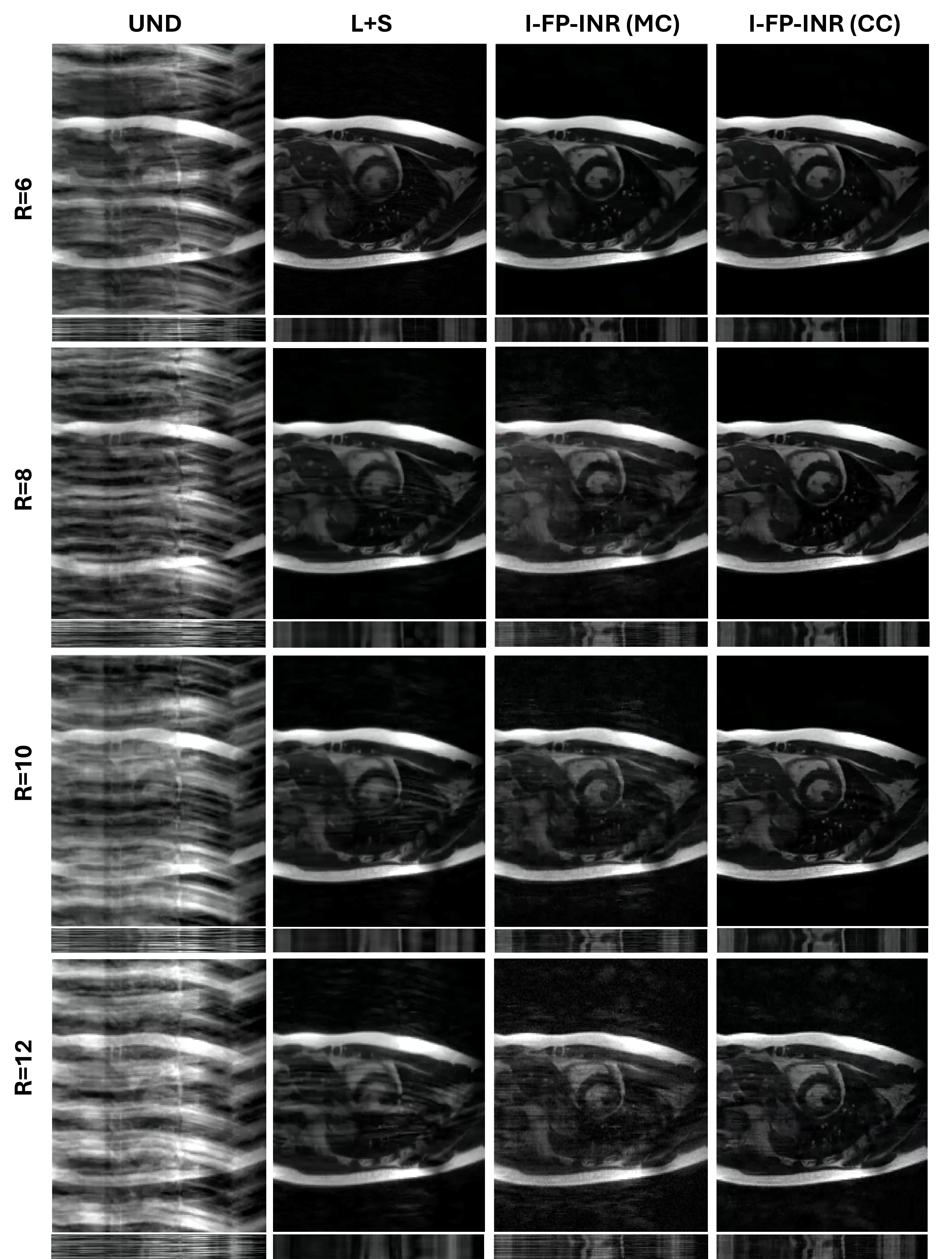}%
\label{fig_lumc_dynamic}}
\caption{Qualitative visualizations of the proposed I-FP-INR on in-house data at different acceleration factors under a variable-density sampling pattern. The bottom panels depict temporal profiles along the central line. (a) shows results with a fixed sampling mask, while (b) presents results using a dynamic mask with doubled acceleration factors.}
\label{fig_f}
\end{figure*}

Fig.~\ref{fig_f} shows visual comparisons under different sampling mask conditions. With a fixed sampling mask, I-FP-INR in both reconstruction strategies outperforms L+S, producing sharper features and better structural recovery, particularly evident in the bottom section showing the temporal dynamics. At lower acceleration factors (R=3 and 4), I-FP-INR (MC) and I-FP-INR (CC) perform similarly, but as acceleration increases, residual artifacts become more clearly observed and the performance gap widens, consistent with the trends observed in the public data. This is because at lower acceleration factors, the central k-space region is largely sampled, supporting stable optimization, while at higher factors, the proportion of sampled central k-space decreases, making the advantage of I-FP-INR (CC) increasingly evident. Under varying masks, similar trends are observed. Although L+S benefits from improved spatio-temporal coverage, its quality degrades at higher acceleration. I-FP-INR (MC) also declines from R=8, as central k-space sampling is reduced despite better global k-space learning. In contrast, I-FP-INR (CC) maintains performance comparable to the fixed-mask case even at doubled acceleration, demonstrating the importance of effectively exploiting spatio-temporal features and robustness of coil-combined strategy. To note, since the varying mask is derived from fixed-mask regions, I-FP-INR (CC) has the potential to achieve even better performance when extended to the whole k-space sampling scenarios.

\section{Discussion}
In this paper, we propose an image-domain dual-branch INR framework for unsupervised cardiac cine MRI reconstruction. By incorporating an additional feature branch to enrich coordinate representations, the method achieved improved performance. Extensive evaluations on two datasets under various sampling scenarios demonstrate its robustness, consistently yielding high-quality reconstruction results.

In the overall comparison with baseline methods, although Hash-INR demonstrates strong performance in its original work, its effectiveness degraded on the testing public dataset. Firstly, Hash-INR employs ReLU activation functions within the INR, which limits its ability to model high-frequency information, as also evidenced by the ablation study in Table~\ref{tab2_ablation1}. Moreover, Hash-INR operates on cropped data centered on the cardiac region in its original operation, thereby reducing reconstruction difficulty by limiting the k-space extent. In addition, a clear distinction is observed between image- and k-space-based INR methods. K-space approaches tend to exhibit blurring and artifacts, as undersampling disrupts the global k-space structure, making fitting more challenging despite the interpolation capability of the INR method. Inconsistencies in the k-space are then readily propagated and manifested as artifacts in the image domain. In contrast, image-domain methods leverage more structured and spatially localized correlations, which are easier for INRs to capture and model, leading to improved reconstruction quality and faster convergence.

Across all sampling patterns, Gaussian Cartesian sampling yielded the best performance, as its fully sampled k-space center lowers the effective acceleration factor. By prioritizing low-frequency regions, it preserves critical structural information and stabilizes the optimization process. In contrast, VISTA pattern without a fully sampled k-space center increases reconstruction difficulty, as the absence of dense low-frequency sampling reduces the strength of data consistency constraints, making recovery of these components more challenging. However, higher central sampling density still captures the dominant image contrast and structural content, supporting reasonable reconstruction. The most challenging setting is the PDS pattern, which enforces more uniform sampling but reduces coverage of the low-frequency region, thereby complicating the learning process and degrading reconstruction quality. Despite this, it can also serve as a rigorous test of a model’s ability to recover structural information when low-frequency guidance is limited.

Another key observation from Table~\ref{tab1} is the performance gap between I-FP-INR (MC) and I-FP-INR (CC) under different sampling patterns. With multi-coil reconstruction strategy, I-FP-INR achieves higher metrics under Gaussian Cartesian sampling with a fully sampled k-space center, where central k-space information supports more reliable exploitation of coil sensitivity and inter-coil redundancy. However, its performance degrades when the fully sampled center is absent, likely due to the increased difficulty in modeling inter-coil correlations and sensitivity variations under sparse sampling. In contrast, I-FP-INR with coil-combined strategy demonstrates more consistent performance across all sampling patterns, highlighting its robustness and better alignment with standard deep learning-based reconstruction pipelines. This suggests that while the multi-coil strategy can better exploit spatial redundancy when sufficient k-space center information is available, the coil-combined one generalizes better under sparse and inconsistent sampling conditions by avoiding the complexities associated with coil sensitivity estimation and inter-coil dependencies.

Although I-FP-INR demonstrates promising performance across various scenarios, it requires longer reconstruction time compared to a conventional INR design that learns direct mappings between coordinates and values. This is mainly due to the inclusion of a feature embedding module, whose parameters need to be continuously updated during optimization to achieve better performance. While the current iteration setting yields good image quality, the number of iterations could potentially be reduced while maintaining comparable performance; however, reconstruction time still remains higher than that of traditional methods. Moreover, observations on in-house data show that the proposed method exhibits residual artifacts under fixed sampling masks at high acceleration factors, where limited k-space coverage hinders effective spatio-temporal modeling. Therefore, improving the efficiency of I-FP-INR and enhancing its robustness in such challenging scenarios represent important directions for future work.

\section{Conclusion}
In this work, we investigate the use of implicit neural representations (INRs) for unsupervised cardiac cine MRI reconstruction. To leverage the benefits of feature embeddings, we introduce an additional INR branch that incorporates extracted image features and interacts with the original coordinate-based branch, resulting in improved reconstruction performance. We further employ an alternating inference stage to progressively refine the input image quality and enhance feature representations. By evaluating I-FP-INR with two coil-format reconstruction strategies, we show that each configuration offers distinct advantages under different sampling patterns. Extensive experiments on both public and in-house data demonstrate that the proposed method achieves strong performance and robust generalization across various vendors, field strengths, sampling patterns and acceleration factors.

\section*{Acknowledgments}
This publication is part of the project ROBUST: Trustworthy AI-based Systems for Sustainable Growth with project number KICH3.LTP.20.006, which is (partly) financed by the Dutch Research Council (NWO), Philips Research, and the Dutch Ministry of Economic Affairs and Climate Policy (EZK) under the program LTP KIC 2020-2023.



 
%


\bibliographystyle{IEEEtran}
\bibliography{ref}

@article{1,
  title={Interventions to alleviate patient anxiety during magnetic resonance imaging: a review},
  author={Phillips, Susan and Deary, Ian J},
  journal={Radiography},
  volume={1},
  number={1},
  pages={29--34},
  year={1995},
  publisher={Elsevier}
}

@article{2,
  title={SENSE: sensitivity encoding for fast MRI},
  author={Pruessmann, Klaas P and Weiger, Markus and Scheidegger, Markus B and Boesiger, Peter},
  journal={Magnetic Resonance in Medicine: An Official Journal of the International Society for Magnetic Resonance in Medicine},
  volume={42},
  number={5},
  pages={952--962},
  year={1999},
  publisher={Wiley Online Library}
}

@article{3,
  title={A Review on Accelerated Magnetic Resonance Imaging Techniques: Parallel Imaging, Compressed Sensing, and Machine Learning},
  author={Tavakkoli, Mitra and Noseworthy, Michael D},
  journal={Critical Reviews™ in Biomedical Engineering},
  volume={53},
  number={5},
  year={2025},
  publisher={Begel House Inc.}
}

@article{4,
  title={MoDL: Model-based deep learning architecture for inverse problems},
  author={Aggarwal, Hemant K and Mani, Merry P and Jacob, Mathews},
  journal={IEEE transactions on medical imaging},
  volume={38},
  number={2},
  pages={394--405},
  year={2018},
  publisher={IEEE}
}

@article{5,
  title={ADMM-CSNet: A deep learning approach for image compressive sensing},
  author={Yang, Yan and Sun, Jian and Li, Huibin and Xu, Zongben},
  journal={IEEE transactions on pattern analysis and machine intelligence},
  volume={42},
  number={3},
  pages={521--538},
  year={2018},
  publisher={IEEE}
}

@article{6,
  title={IDPCNN: Iterative denoising and projecting CNN for MRI reconstruction},
  author={Hou, Ruizhi and Li, Fang},
  journal={Journal of Computational and Applied Mathematics},
  volume={406},
  pages={113973},
  year={2022},
  publisher={Elsevier}
}

@article{7,
  title={An adaptive intelligence algorithm for undersampled knee MRI reconstruction},
  author={Pezzotti, Nicola and Yousefi, Sahar and Elmahdy, Mohamed S and Van Gemert, Jeroen Hendrikus Fransiscus and Schuelke, Christophe and Doneva, Mariya and Nielsen, Tim and Kastryulin, Sergey and Lelieveldt, Boudewijn PF and Van Osch, Matthias JP and others},
  journal={Ieee Access},
  volume={8},
  pages={204825--204838},
  year={2020},
  publisher={IEEE}
}

@article{8,
  title={Score-based diffusion models for accelerated MRI},
  author={Chung, Hyungjin and Ye, Jong Chul},
  journal={Medical image analysis},
  volume={80},
  pages={102479},
  year={2022},
  publisher={Elsevier}
}

@article{9,
  title={Convolutional recurrent neural networks for dynamic MR image reconstruction},
  author={Qin, Chen and Schlemper, Jo and Caballero, Jose and Price, Anthony N and Hajnal, Joseph V and Rueckert, Daniel},
  journal={IEEE transactions on medical imaging},
  volume={38},
  number={1},
  pages={280--290},
  year={2018},
  publisher={IEEE}
}

@article{10,
  title={Convolutional recurrent U-net for cardiac cine MRI reconstruction via effective spatio-temporal feature exploitation},
  author={Lyu, Donghang and Staring, Marius and van Osch, Matthias JP and Doneva, Mariya and Lamb, Hildo J and Pezzotti, Nicola},
  journal={Medical Physics},
  volume={53},
  number={1},
  pages={e70245},
  year={2026},
  publisher={Wiley Online Library}
}

@inproceedings{11,
  title={Fill the k-space and refine the image: Prompting for dynamic and multi-contrast MRI reconstruction},
  author={Xin, Bingyu and Ye, Meng and Axel, Leon and Metaxas, Dimitris N},
  booktitle={International Workshop on Statistical Atlases and Computational Models of the Heart},
  pages={261--273},
  year={2023},
  organization={Springer}
}

@article{12,
  title={Self-supervised learning of physics-guided reconstruction neural networks without fully sampled reference data},
  author={Yaman, Burhaneddin and Hosseini, Seyed Amir Hossein and Moeller, Steen and Ellermann, Jutta and U{\u{g}}urbil, K{\^a}mil and Ak{\c{c}}akaya, Mehmet},
  journal={Magnetic resonance in medicine},
  volume={84},
  number={6},
  pages={3172--3191},
  year={2020},
  publisher={Wiley Online Library}
}

@article{13,
  title={Clean self-supervised MRI reconstruction from noisy, sub-sampled training data with Robust SSDU},
  author={Millard, Charles and Chiew, Mark},
  journal={Bioengineering},
  volume={11},
  number={12},
  pages={1305},
  year={2024},
  publisher={MDPI}
}

@inproceedings{14,
  title={Neural implicit k-space for binning-free non-cartesian cardiac MR imaging},
  author={Huang, Wenqi and Li, Hongwei Bran and Pan, Jiazhen and Cruz, Gastao and Rueckert, Daniel and Hammernik, Kerstin},
  booktitle={International Conference on Information Processing in Medical Imaging},
  pages={548--560},
  year={2023},
  organization={Springer}
}

@article{15,
  title={Multi-dynamic deep image prior for cardiac MRI},
  author={Vornehm, Marc and Chen, Chong and Sultan, Muhammad A and Arshad, Syed M and Han, Yuchi and Knoll, Florian and Ahmad, Rizwan},
  journal={Magnetic Resonance in Medicine},
  volume={94},
  number={6},
  pages={2668--2679},
  year={2025},
  publisher={Wiley Online Library}
}

@article{16,
  title={Noise2Noise: Learning image restoration without clean data},
  author={Lehtinen, Jaakko and Munkberg, Jacob and Hasselgren, Jon and Laine, Samuli and Karras, Tero and Aittala, Miika and Aila, Timo},
  journal={arXiv preprint arXiv:1803.04189},
  year={2018}
}

@article{17,
  title={A theoretical framework for self-supervised MR image reconstruction using sub-sampling via variable density Noisier2Noise},
  author={Millard, Charles and Chiew, Mark},
  journal={IEEE transactions on computational imaging},
  volume={9},
  pages={707--720},
  year={2023},
  publisher={IEEE}
}

@inproceedings{18,
  title={Deep image prior},
  author={Ulyanov, Dmitry and Vedaldi, Andrea and Lempitsky, Victor},
  booktitle={Proceedings of the IEEE conference on computer vision and pattern recognition},
  pages={9446--9454},
  year={2018}
}

@article{19,
  title={Time-dependent deep image prior for dynamic MRI},
  author={Yoo, Jaejun and Jin, Kyong Hwan and Gupta, Harshit and Yerly, Jerome and Stuber, Matthias and Unser, Michael},
  journal={IEEE Transactions on Medical Imaging},
  volume={40},
  number={12},
  pages={3337--3348},
  year={2021},
  publisher={IEEE}
}

@inproceedings{20,
  title={Implicit functions in feature space for 3d shape reconstruction and completion},
  author={Chibane, Julian and Alldieck, Thiemo and Pons-Moll, Gerard},
  booktitle={Proceedings of the IEEE/CVF conference on computer vision and pattern recognition},
  pages={6970--6981},
  year={2020}
}

@inproceedings{22,
  title={Implicit neural representations for deformable image registration},
  author={Wolterink, Jelmer M and Zwienenberg, Jesse C and Brune, Christoph},
  booktitle={International Conference on medical imaging with deep learning},
  pages={1349--1359},
  year={2022},
  organization={PMLR}
}

@inproceedings{23,
  title={Cina: Conditional implicit neural atlas for spatio-temporal representation of fetal brains},
  author={Dannecker, Maik and Kyriakopoulou, Vanessa and Cordero-Grande, Lucilio and Price, Anthony N and Hajnal, Joseph V and Rueckert, Daniel},
  booktitle={International Conference on Medical Image Computing and Computer-Assisted Intervention},
  pages={181--191},
  year={2024},
  organization={Springer}
}

@article{24,
  title={Spatiotemporal implicit neural representation for unsupervised dynamic MRI reconstruction},
  author={Feng, Jie and Feng, Ruimin and Wu, Qing and Shen, Xin and Chen, Lixuan and Li, Xin and Feng, Li and Chen, Jingjia and Zhang, Zhiyong and Liu, Chunlei and others},
  journal={IEEE Transactions on Medical Imaging},
  volume={44},
  number={5},
  pages={2143--2156},
  year={2025},
  publisher={IEEE}
}

@article{25,
  title={Implicit neural networks with fourier-feature inputs for free-breathing cardiac MRI reconstruction},
  author={Kunz, Johannes F and Ruschke, Stefan and Heckel, Reinhard},
  journal={IEEE Transactions on Computational Imaging},
  volume={10},
  pages={1280--1289},
  year={2024},
  publisher={IEEE}
}

@article{26,
  title={PISCO: Self-supervised k-space regularization for improved neural implicit k-space representations of dynamic MRI},
  author={Spieker, Veronika and Eichhorn, Hannah and Huang, Wenqi and Stelter, Jonathan K and Catalan, Tabita and Braren, Rickmer F and Rueckert, Daniel and Costabal, Francisco Sahli and Hammernik, Kerstin and Karampinos, Dimitrios C and others},
  journal={Medical Image Analysis},
  pages={103890},
  year={2025},
  publisher={Elsevier}
}

@inproceedings{27,
  title={KP-INR: A Dual-Branch Implicit Neural Representation Model for Cardiac Cine MRI Reconstruction},
  author={Lyu, Donghang and Staring, Marius and Doneva, Mariya and Lamb, Hildo J and Pezzotti, Nicola},
  booktitle={International Workshop on Statistical Atlases and Computational Models of the Heart},
  pages={56--66},
  year={2025},
  organization={Springer}
}

@inproceedings{28,
  title={Implicit functions in feature space for 3d shape reconstruction and completion},
  author={Chibane, Julian and Alldieck, Thiemo and Pons-Moll, Gerard},
  booktitle={Proceedings of the IEEE/CVF conference on computer vision and pattern recognition},
  pages={6970--6981},
  year={2020}
}

@article{29,
  title={CMRxRecon2024: a multimodality, multiview k-space dataset boosting universal machine learning for accelerated cardiac MRI},
  author={Wang, Zi and Wang, Fanwen and Qin, Chen and Lyu, Jun and Ouyang, Cheng and Wang, Shuo and Li, Yan and Yu, Mengyao and Zhang, Haoyu and Guo, Kunyuan and others},
  journal={Radiology: Artificial Intelligence},
  volume={7},
  number={2},
  pages={e240443},
  year={2025},
  publisher={Radiological Society of North America}
}

@article{30,
  title={Accelerated dynamic MRI exploiting sparsity and low-rank structure: kt SLR},
  author={Lingala, Sajan Goud and Hu, Yue and DiBella, Edward and Jacob, Mathews},
  journal={IEEE transactions on medical imaging},
  volume={30},
  number={5},
  pages={1042--1054},
  year={2011},
  publisher={IEEE}
}

@article{31,
  title={Image reconstruction from highly undersampled (k, t)-space data with joint partial separability and sparsity constraints},
  author={Zhao, Bo and Haldar, Justin P and Christodoulou, Anthony G and Liang, Zhi-Pei},
  journal={IEEE transactions on medical imaging},
  volume={31},
  number={9},
  pages={1809--1820},
  year={2012},
  publisher={IEEE}
}

@article{32,
  title={Nerf: Representing scenes as neural radiance fields for view synthesis},
  author={Mildenhall, Ben and Srinivasan, Pratul P and Tancik, Matthew and Barron, Jonathan T and Ramamoorthi, Ravi and Ng, Ren},
  journal={Communications of the ACM},
  volume={65},
  number={1},
  pages={99--106},
  year={2021},
  publisher={ACM New York, NY, USA}
}

@article{33,
  title={Fourier features let networks learn high frequency functions in low dimensional domains},
  author={Tancik, Matthew and Srinivasan, Pratul and Mildenhall, Ben and Fridovich-Keil, Sara and Raghavan, Nithin and Singhal, Utkarsh and Ramamoorthi, Ravi and Barron, Jonathan and Ng, Ren},
  journal={Advances in neural information processing systems},
  volume={33},
  pages={7537--7547},
  year={2020}
}

@inproceedings{34,
  title={Wire: Wavelet implicit neural representations},
  author={Saragadam, Vishwanath and LeJeune, Daniel and Tan, Jasper and Balakrishnan, Guha and Veeraraghavan, Ashok and Baraniuk, Richard G},
  booktitle={Proceedings of the IEEE/CVF conference on computer vision and pattern recognition},
  pages={18507--18516},
  year={2023}
}

@article{35,
  title={Image quality assessment: Unifying structure and texture similarity},
  author={Ding, Keyan and Ma, Kede and Wang, Shiqi and Simoncelli, Eero P},
  journal={IEEE transactions on pattern analysis and machine intelligence},
  volume={44},
  number={5},
  pages={2567--2581},
  year={2020},
  publisher={IEEE}
}

@article{36,
  title={A Haar wavelet-based perceptual similarity index for image quality assessment},
  author={Reisenhofer, Rafael and Bosse, Sebastian and Kutyniok, Gitta and Wiegand, Thomas},
  journal={Signal Processing: Image Communication},
  volume={61},
  pages={33--43},
  year={2018},
  publisher={Elsevier}
}

@article{37,
  title={Image quality assessment for magnetic resonance imaging},
  author={Kastryulin, Sergey and Zakirov, Jamil and Pezzotti, Nicola and Dylov, Dmitry V},
  journal={IEEE Access},
  volume={11},
  pages={14154--14168},
  year={2023},
  publisher={IEEE}
}

@article{38,
  title={Low-rank plus sparse matrix decomposition for accelerated dynamic MRI with separation of background and dynamic components},
  author={Otazo, Ricardo and Candes, Emmanuel and Sodickson, Daniel K},
  journal={Magnetic resonance in medicine},
  volume={73},
  number={3},
  pages={1125--1136},
  year={2015},
  publisher={Wiley Online Library}
}

@inproceedings{39,
  title={Practical parallel imaging compressed sensing MRI: Summary of two years of experience in accelerating body MRI of pediatric patients},
  author={Vasanawala, Shreyas S and Murphy, MJ and Alley, Marcus T and Lai, Peng and Keutzer, Kurt and Pauly, John M and Lustig, Michael},
  booktitle={2011 ieee international symposium on biomedical imaging: From nano to macro},
  pages={1039--1043},
  year={2011},
  organization={IEEE}
}

@article{40,
  title={Variable density incoherent spatiotemporal acquisition (VISTA) for highly accelerated cardiac MRI},
  author={Ahmad, Rizwan and Xue, Hui and Giri, Shivraman and Ding, Yu and Craft, Jason and Simonetti, Orlando P},
  journal={Magnetic resonance in medicine},
  volume={74},
  number={5},
  pages={1266--1278},
  year={2015},
  publisher={Wiley Online Library}
}

@article{41,
  title={Implicit neural representations with periodic activation functions},
  author={Sitzmann, Vincent and Martel, Julien and Bergman, Alexander and Lindell, David and Wetzstein, Gordon},
  journal={Advances in neural information processing systems},
  volume={33},
  pages={7462--7473},
  year={2020}
}

@article{42,
  title={Generalized autocalibrating partially parallel acquisitions (GRAPPA)},
  author={Griswold, Mark A and Jakob, Peter M and Heidemann, Robin M and Nittka, Mathias and Jellus, Vladimir and Wang, Jianmin and Kiefer, Berthold and Haase, Axel},
  journal={Magnetic Resonance in Medicine: An Official Journal of the International Society for Magnetic Resonance in Medicine},
  volume={47},
  number={6},
  pages={1202--1210},
  year={2002},
  publisher={Wiley Online Library}
}

@article{43,
  title={Sparse MRI: The application of compressed sensing for rapid MR imaging},
  author={Lustig, Michael and Donoho, David and Pauly, John M},
  journal={Magnetic Resonance in Medicine: An Official Journal of the International Society for Magnetic Resonance in Medicine},
  volume={58},
  number={6},
  pages={1182--1195},
  year={2007},
  publisher={Wiley Online Library}
}

@article{44,
  title={Electrodynamics and ultimate SNR in parallel MR imaging},
  author={Wiesinger, Florian and Boesiger, Peter and Pruessmann, Klaas P},
  journal={Magnetic Resonance in Medicine: An Official Journal of the International Society for Magnetic Resonance in Medicine},
  volume={52},
  number={2},
  pages={376--390},
  year={2004},
  publisher={Wiley Online Library}
}

@inproceedings{45,
  title={Vestibular schwannoma growth prediction from longitudinal MRI by time-conditioned neural fields},
  author={Chen, Yunjie and Wolterink, Jelmer M and Neve, Olaf M and Romeijn, Stephan R and Verbist, Berit M and Hensen, Erik F and Tao, Qian and Staring, Marius},
  booktitle={International Conference on Medical Image Computing and Computer-Assisted Intervention},
  pages={508--518},
  year={2024},
  organization={Springer}
}

\end{document}